\documentclass[conference]{IEEEtran}
\usepackage{cite}
\usepackage{nameref}

\newif\ifarxiv
\arxivfalse
\ifarxiv \PassOptionsToPackage{draft}{graphicx}\fi
\usepackage{graphicx}
\usepackage{enumitem}
\usepackage{tikz}
\usepackage[edges]{forest}
\usepackage{adjustbox}
\usepackage{color}
\usepackage{subcaption}
\usepackage{hyperref}
\usepackage{booktabs}
\usepackage{longtable}
\usepackage{multicol}
\usepackage{multirow}
\usepackage{enumitem}
\usepackage{cite}
\usepackage{nameref}
\usepackage{amsmath}
\usepackage{soul}
\usepackage{changepage}
\usepackage{fontawesome5} 
\usepackage{svg}
\usepackage{amsmath}
\usepackage{makecell} 

\def\BibTeX{{\rm B\kern-.05em{\sc i\kern-.025em b}\kern-.08em
    T\kern-.1667em\lower.7ex\hbox{E}\kern-.125emX}}

\definecolor{visnecurugu}{rgb}{0.6,0.1,0.6}
\definecolor{heredot}{RGB}{100, 100, 255}
\definecolor{byzantium}{RGB}{112, 41, 99}

\usepackage[normalem]{ulem}

\usetikzlibrary{arrows.meta, shapes.geometric, positioning, backgrounds}

\title{UAV-SEAD: State Estimation Anomaly Dataset for UAVs}

\author{Aykut Kabaoglu$^{1}$, Sanem Sariel$^{2}$}

\date{February, 2026}

\begin{document}

\maketitle
\footnotetext[1]{Artificial Intelligence and Robotics Laboratory (AIR Lab), Department of Computer Engineering, Istanbul Technical University, Istanbul, Türkiye. kabaoglua@itu.edu.tr}
\footnotetext[2]{Artificial Intelligence and Robotics Laboratory (AIR Lab), Faculty of Computer and Informatics Engineering, Istanbul Technical University, Istanbul, Türkiye. sariel@itu.edu.tr}

\begin{abstract}
Accurate state estimation in Unmanned Aerial Vehicles (UAVs) is crucial for ensuring reliable and safe operation, as anomalies occurring during mission execution may induce discrepancies between expected and observed system behaviors, thereby compromising mission success or posing potential safety hazards. It is essential to continuously monitor and detect such conditions in order to ensure a timely response and maintain system reliability. In this work, we focus on UAV state estimation anomalies and provide a large-scale real-world UAV dataset to facilitate research aimed at improving the development of anomaly detection. Unlike existing datasets that primarily rely on injected faults into simulated data, this dataset comprises 1396 real flight logs totaling over 52 hours of flight time, collected across diverse indoor and outdoor environments using a collection of PX4-based UAVs equipped with a variety of sensor configurations. The dataset comprises both normal and anomalous flights without synthetic manipulation, making it uniquely suitable for realistic anomaly detection tasks. A structured classification is proposed that categorizes UAV state estimation anomalies into four classes: mechanical and electrical, external position, global position, and altitude anomalies. These classifications reflect collective, contextual, and outlier anomalies observed in multivariate sensor data streams, including IMU, GPS, barometer, magnetometer, distance sensors, visual odometry, and optical flow, that can be found in the PX4 logging mechanism. It is anticipated that this dataset will play a key role in the development, training, and evaluation of anomaly detection and isolation systems to address the critical gap in UAV reliability research.

\end{abstract}


\section{Introduction} \label{introduction}

Unmanned and autonomous vehicles are playing an increasingly vital role, with their use growing across a wide range of domains. Unmanned Aerial Vehicles (UAVs) are among the most safety-critical systems because of their operational nature. They generally operate within constrained environments, and require a high degree of stability and robustness; otherwise, they may face a significant risk of crashing or failing \cite{banerjee2020risk}. Even minor anomalies can cause catastrophic failures, which pose significant dangers to their operating environments. Based on this premise, this study focuses on the classification of state estimation anomalies in UAVs and presents a dataset to support the development of anomaly detection and prediction methods.

Establishing a precise taxonomy of faults and failures is a prerequisite for understanding the nature of anomalies, as these concepts represent the sequential stages from the root cause of a defect to its ultimate impact on system integrity.
A fault is formally defined as any unpermitted condition of at least one characteristic property of the system from standard behavior \cite{keipour2019automatic} and may manifest as broken or malfunctioning components or subsystems. A fault may exist without immediately leading to system collapse, especially if control systems are robust \cite{freeman2013model}.
A failure is the complete interruption of a system’s ability to perform a required function under specified operating conditions \cite{puchalski2022uav}  or represents a disruption in the intended functionality of the system, where one or more components stop operating as designed, resulting in a performance breakdown \cite{gupta2018open}. Therefore, a failure is the final, definitive event where the system is permanently unable to perform its function \cite{puchalski2022uav}. 
In contrast, an anomaly is a deviation from the expected behavior of the system that can indicate underlying faults or emerging failures, often requiring detection before compromising performance or safety \cite{khalastchi2018fault}. It does not necessarily mean that the system has failed; rather, it indicates that some aspect of its performance or state is outside of the predefined or anticipated working conditions.
In summary, faults are defects in a system property, anomalies are deviations in behavior, and failures are the resulting loss of system functionality. While anomalies can serve as early indicators of potential failures, not all anomalies necessarily result in failure.

The quality and design of the hardware and software systems influence the likelihood that anomalies occur in monitored vehicles. Employing better hardware or increasing the number of measurement sources may increase the system quality; however, hardware costs are often a factor in the development process, and no hardware is completely error-free. Limiting the operating environment can reduce the probability of anomalies by controlling the working conditions. Additionally, modeling the system allows us to predict the expected behavior of the vehicle and its environment. However, this approach possibly cannot cover all potential anomalies, and can be expensive to build. 

All of these factors introduce uncertainties for both the vehicle and its surrounding environment. These uncertainties can arise from incomplete information, unreliable resources, the stochastic nature of complex devices, environmental variables, unobservable conditions, limited resources, abstract knowledge, source noise, and representation of the selected source for the current environment \cite{pettersson2005execution}. Although completely preventing uncertainties may be preferable, it is not always possible. Therefore, managing them is crucial for real-world applications in order to build robust systems that are resilient to potential failures.
To address this need, monitoring systems are designed and implemented to observe the system behavior. These systems can function as loggers for future analysis or as fault detection and isolation (FDI) systems. Fault detection identifies when something is going wrong in the monitored system, while fault isolation classifies what specifically is malfunctioning \cite{pettersson2005execution}. Additionally, the fault’s magnitude is determined by fault identification. Often, the term fault diagnosis is used synonymously with fault isolation. FDI systems help prevent failures through early detection or initiate recovery after a failure has occurred.

In unmanned vehicles, especially UAVs, anomalies can arise from hardware (e.g., sensors, actuators, computational units, power fluctuations, mechanical structure), software (e.g., mapping, localization, planning, filtering, decision-making, controllers, actions), and the environment (e.g., natural forces, unexpected impacts, sudden changes, interruptions, illumination). Autonomous vehicles must compute their state, such as location on a map or position relative to a reference, by combining orientation data to navigate without remote control. The state (position/orientation) of an unmanned vehicle can be affected by any of the aforementioned anomaly sources, and detecting any anomalies is crucial to completing a desired mission or execution on the vehicle.

UAVs require robust systems to maintain state stability. State estimation anomalies represent the divergence between the ground-truth position/orientation and the estimated position/orientation output. When an anomaly is detected before a failure occurs, failsafe or recovery actions can be triggered on the UAV. 

There are three main approaches in the literature to detect and classify anomalies: model-based, knowledge-based, and data-driven \cite{kang2025real}. Model-based approaches construct mathematical or analytical representations of a system’s normal behavior, such as physical models or state-space models. Anomalies are identified when the observed data deviate significantly from the predictions of these models. Knowledge-based approaches, sometimes referred to as rule-based or expert systems, rely on domain expertise to define conditions that characterize abnormal behavior. These rules may be expressed through logical constraints, fuzzy inference systems, or heuristic thresholds. In contrast, data-driven approaches employ statistical analysis and machine learning techniques to infer patterns of normal and abnormal behavior directly from the data. Instead of relying on explicit rules or physical models, they learn baselines from historical observations and detect deviations, or in some cases, they directly classify anomalous states.

Data-driven anomaly detection systems can effectively learn the behavior of the investigated systems and identify unexpected situations, provided system has sufficient capability to represent both the working environment and the UAV. However, collecting enough data to replicate all possible system behaviors is challenging. The real-world data collection process can be time-consuming. Creating a dataset for UAVs is more effort intensive than for ground vehicles due to government regulations and the difficulty of finding suitable flight conditions. Moreover, replicating an anomaly in real conditions can be dangerous and may result in accidents. Finding a natural anomaly is a more complex task than creating an environment to replicate one, as hundreds of flights may be required to naturally encounter a single anomaly.

The state estimation anomaly refers to an anomalous behavior observable in the estimated state variables, regardless of whether its root cause originates from sensors, the estimator, or the environment. In this work, we first present a structured classification covering the possible categories of state estimation anomalies encountered in UAV operations, then introduce UAV-SEAD, a dataset for UAV state estimation anomalies that combines data collected from real flight experiments performed in indoor and outdoor settings. UAV-SEAD is the largest UAV flight dataset in the literature so far for anomaly detection, considering the number of real flights.

The remainder of this paper is structured as follows: Section \ref{related_works} reviews related work on UAV anomaly detection, Section \ref{state_estimation_anomalies} introduces the state estimation anomaly problem in UAVs, Section \ref{working_environment} describes the data collection environment and hardware, Section \ref{anomaly_classification} presents state estimation anomaly classifications for UAVs, Section \ref{uav_sead_dataset} provides an evaluation of the performed flights, Section \ref{comparison_of_datasets} compares UAV anomaly datasets, Section \ref{conclusion_and_future_works} concludes the study and outlines future research directions.


\section{Related Works} \label{related_works}

Fault and anomaly detection and identification research encompasses diverse domains, ranging from structural health monitoring and communication link analysis to the detection of propulsion system failures and anomalous sensor-state behaviors. This section primarily focuses on research concerning robots and autonomous vehicles, with a particular emphasis on Unmanned Aerial Vehicles (UAVs). While these platforms present unique and relatively underexplored challenges, various established methodologies have been adapted to address their specific requirements. In general, existing literature can be categorized by methodology into three main approaches: model-based, knowledge-based, and data-driven anomaly detection.

Model-based approaches generally operate on the assumption that significant deviations between model estimations and actual sensor measurements indicate the presence of faults. For instance, Venkataraman et al. \cite{venkataraman2019comparison} evaluate three fault detection and isolation (FDI) methods for addressing stuck elevon faults in fixed-wing aircraft. These methods include parity checks, which compare sensor outputs against model-generated outputs using predefined thresholds, as well as observer-based models for linear parameter-varying systems and observer-based multi-model adaptive estimators.
A widely utilized approach in fault detection involves generating residuals, which represent the discrepancies between estimated and measured system states. While residuals are highly representative of system health, they are often sensitive to measurement noise. Common residual generation techniques include parity equations, state observers, and parameter estimation methods. Furthermore, Freeman et al. \cite{freeman2013model} emphasize that a reliable FDI system requires a residual generator with several key characteristics: sensitivity to specific faults while remaining invariant to others, robustness against modeling uncertainties, and the effective suppression of external disturbances and noise. Additionally, an ideal generator must be capable of differentiating between simultaneous faults and maintaining a non-zero steady-state output in the presence of a fault while returning to zero otherwise.
At the same time, Khalastchi et al. \cite{khalastchi2013sensor} highlight two particularly challenging fault types: sensor stuck and drift. A 'stuck' fault occurs when a sensor maintains a constant output regardless of changes in the actual state, while 'drift' involves a gradual deviation of sensor readings from the true value over time. To address these issues, they utilize an Unscented Kalman Filter (UKF) integrated with a nonlinear model of aircraft dynamics. This approach enables robust state estimation for UAVs equipped with multiple IMUs, creating a system resilient to individual sensor failures.
Additionally, Keipour et al. \cite{keipour2019automatic} propose a real-time anomaly detection framework for fixed-wing UAVs that utilizes Recursive Least Squares (RLS) to model input-output behavior online. By flagging deviations from this learned behavior, the system was validated using the ALFA dataset \cite{keipour2021alfa}.
Knowledge-based approaches, often categorized as expert or rule-based systems, rely on domain expertise to establish logical constraints and heuristic thresholds that characterize abnormal behavior. Unlike purely mathematical models, these methods prioritize the semantic analysis of vehicle dynamics to identify physically contradictory states. For instance, Islam et al. \cite{islam2024adam} introduces the Adaptive Drone Anomaly Monitor (ADAM), which enhances runtime reliability by integrating sensor innovations and control outputs with environmental data to detect flight deviations.

While model-based methods rely on accurate representations of aircraft or robot dynamics, they often struggle with small-scale UAVs due to incomplete modeling and unmodeled disturbances. To address these limitations, hybrid techniques integrate model-based residuals with machine learning algorithms. For example, Bronz et al. \cite{bronz2020real} propose a real-time fault detector for small fixed-wing UAVs that utilizes sensor residuals in conjunction with lightweight machine learning classifiers to adaptively distinguish between normal and faulty actuator behaviors.
Sun et al. \cite{sun2017novel} present an online, data-driven algorithm that combines Kalman filter residuals with an adaptive neuro-fuzzy inference system (ANFIS) to detect navigation sensor faults on UAVs. By continuously monitoring residuals from a real-time Kalman filter, the method applies neuro-fuzzy logic to classify anomalies into point, contextual, and collective faults. This approach facilitates rapid, in-flight fault identification by leveraging the learning capabilities of neural networks alongside the rule-based reasoning of fuzzy logic.
For autonomous systems, fault detection alone is insufficient; faults must also be identified to enable appropriate response or recovery maneuvers. Goel et al. \cite{goel2000fault} address this by proposing a hybrid fault detection and identification (FDI) framework that presents a neural network architecture integrating the outputs from multiple models for accurate classification. To enhance identification precision in wheeled mobile robots, their approach assigns dedicated Kalman filters to monitor specific failure modes, such as those involving encoders, tires, or gyroscopes.

Data-driven techniques bypass predefined system models by leveraging large volumes of labeled or unlabeled flight data to learn normal versus anomalous behavior. In addition to traditional methodologies, machine learning has become a prominent topic in recent studies, offering promising solutions for anomaly and fault detection, especially after the increasing accessibility of datasets, and data-driven approaches are better suited to modeling the complex, highly nonlinear systems.

Many existing algorithms detect faults by identifying anomalies in individual attribute values while ignoring their contextual relationships. Contextual faults occur when a defective sensor produces values that are invalid with respect to other attributes. Alos and Dahrouj \cite{alos2020decision} introduce a novel matrix platform for detecting potential contextual faults; this platform consists of multiple small decision trees instead of a single, large decision tree, which would be difficult and time-consuming to construct for datasets with high dimensionality.
Moreover, Titouna et al. \cite{titouna2020online} propose an online anomaly detection framework for UAV fleets that utilizes a two-step approach: it first computes Kullback–Leibler divergence on external sensor distributions among neighboring vehicles to flag spatial outliers. It then employs a Multi-Layer Perceptron (MLP) trained on internal sensor readings (e.g., temperature, battery level) to isolate the physical failure within the identified faulty UAV.
For instance, Dhakal et al. \cite{dhakal2023uav} propose unsupervised learning techniques to learn the behavior of UAVs from normal flights by using Autoencoders (AE) and Variational Autoencoders (VAE). In this framework, anomaly detection is achieved by establishing thresholds based on reconstruction errors, where significant deviations from the learned normal state signify a fault. Furthermore, the authors provide a comparative analysis of the performance metrics between AE and VAE architectures to evaluate their respective detection efficiencies.

Since UAV sensor data are inherently multivariate time series, they are well-suited for sequential machine learning architectures and time-series forecasting models.
In this context, Temporal Convolutional Networks (TCNs) have gained significant attention as a robust data-driven approach. As a general framework for dynamic time-series modeling, TCNs offer broad receptive fields and parallel processing capabilities, providing a versatile convolutional neural network architecture for addressing complex temporal dependencies \cite{you2022adaptable}.
Recent research emphasizes the importance of capturing both inter-sensor dependencies and temporal dynamics through deep architectures. Deng and Hooi \cite{deng2021graph} introduced the Graph Deviation Network (GDN), a framework that constructs a directed sensor graph by learning pairwise dependencies from nominal data. By employing message-passing and attention mechanisms, GDN captures dynamic correlations within multivariate time series. Modeling each sensor as a node and leveraging learned edge weights allows the system to detect anomalies manifesting as unexpected shifts in inter-sensor relationships, moving beyond the limitations of simple univariate thresholding.

Many studies in the literature often focus on actuator, sensor, navigation, or battery faults, and mostly rely on predefined parameters determined by observers. While Kalman filters are commonly used for anomaly detection, they represent a purely stochastic approach that combines the covariances of each data source over time and do not inherently account for the specific vehicle type or its attributes. A significant limitation is that even model-based approaches often fail to adequately consider vehicle dynamics. Furthermore, the process of accurately measuring the covariances of each input is challenging, as these values can change over time and are highly dependent on calibration. To compound this issue, these values can vary by hardware, and manufacturers often only provide average values, if they provide them at all. Much of the existing research focuses solely on the vehicle's current state, validating it at a given moment without tracking time intervals to determine if the vehicle could feasibly reach that state. While using a single filtering mechanism like a Bayesian filter for all vehicle types simplifies implementation, it also increases vulnerability by neglecting a vehicle's specific structure or model-specific considerations.

Data-driven methods are emerging approaches to overcome the difficulties of modeling aircraft dynamics and creating robust filtering systems. These methods consider the cumulative effects of aircraft dynamics, hardware variations, and failure conditions by analyzing large volumes of data. However, collecting a substantial amount of flight data required to build a comprehensive anomaly and failure dataset remains difficult. Many existing datasets are limited in scope, often relying on simulated data or containing only a small number of real-world flight records. This paper addresses this critical gap by introducing a large-scale dataset comprising over 52 hours of real-world flights, specifically designed to support the development of UAV anomaly detection systems.


\section{State Estimation Anomalies in UAVs} \label{state_estimation_anomalies}
The state of an unmanned vehicle is defined as a state vector $(\mathbf{s})$ in three-dimensional state space comprising the position $(\vec{\mathbf{p}})$, orientation in quaternion $(\vec{\mathbf{q}})$, and linear velocity $(\vec{\mathbf{v}})$ as given in Equation \ref{state_vector}. 

\begin{equation}
\label{state_vector}
\begin{aligned}
    \mathbf{s} &= < \vec{\mathbf{p}} \hspace{2mm} \vec{\mathbf{q}} \hspace{2mm} \vec{\mathbf{v}} >, \\
    \vec{\mathbf{p}} &= \begin{bmatrix} x & y & z \end{bmatrix}, \\
    \vec{\mathbf{q}} &= \begin{bmatrix} q_x & q_y & q_z & q_w \end{bmatrix}, \\
    \vec{\mathbf{v}} &= \begin{bmatrix} v_x & v_y & v_z \end{bmatrix}, \\
\end{aligned}
\end{equation}

\begin{figure}
    \centering
    \includegraphics[width=\linewidth]{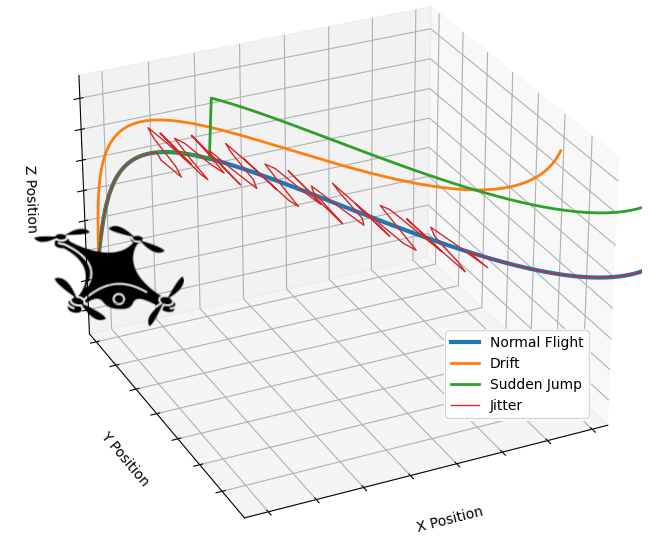}
    \caption{State estimation anomalies as drift, jump, and jitter }
    \label{anomaly_fligth}
\end{figure}

This vector is estimated and maintained by a filtering process and should be continuously monitored against the possibility of state estimation anomalies.
EKF is a commonly used filter for state estimation due to its high performance and low computational cost, which ensures stable performance in most cases. However, it can still propagate incorrect data into the estimated state. Additionally, EKF relies on the given covariances of the input data sources.
Although EKF uses covariances to manage uncertainties and generate more stable outcomes, one of its main stability issues is divergence. Divergence occurs when the covariance matrix of the EKF’s estimated state grows uncontrollably over time, leading to unreliable and incorrect estimates \cite{schlee1967divergence, simon2006optimal}. This can happen for several reasons, such as:

\begin{itemize}
\item Linearization Errors: The EKF linearizes the system dynamics and measurement equations around the current estimated state. If the true system dynamics are highly nonlinear, this linearization can introduce significant errors, especially if the estimated operating point is far from the true state.
\item Incorrect Process or Measurement Models: If the system dynamics or measurement models do not accurately represent the system's true behavior, the EKF may fail to provide accurate estimates.
\item High Noise Levels: High noise levels in the system or measurements can lead to inaccuracies in the estimated state and the covariance matrix, potentially causing instabilities.
\item Poor Initial Estimates: Like the traditional Kalman Filter, the performance of EKF depends on the quality of the initial state estimate and covariance matrix. Poor initial estimates can lead to instabilities.
\item Large Update Time Steps: The EKF assumes that the system evolves smoothly over time, and the prediction relies on a small time step. If the time steps are too large, the linearization assumptions may break down, causing instabilities.
\item Inaccurate Covariance Propagation: During the prediction step, the covariance matrix is propagated using linearized system dynamics. If this propagation is not accurate, it can lead to instabilities.
\end{itemize}

These limitations can be observed as anomaly. In addition to these limitations of EKFs, state estimation anomalies can arise in various ways, influenced by environmental disturbances, system failures, sensor noise, etc. These anomalies may be observed as drift, jitter, or sudden jumps in position or orientation, as shown in Figure \ref{anomaly_fligth}. Gradual deviation of the UAV's position, altitude, or heading from its expected behavior over time is identified as drift in state estimation and can be caused by sensor bias, calibration errors, GPS drift, or some environmental factors like wind, interference, etc. Jitter in state estimation represents rapid, small oscillations or fluctuations in the UAV's position or orientation, and the reasons can be sensor noise, instability, or poor tuning of the control system. Lastly, sudden jumps characterize abrupt changes in the UAV's position caused by GPS glitches, IMU/Barometer measurement errors, SLAM algorithm errors, etc. Also, external position sources may lead to all the anomalous behaviors mentioned above. The timely detection of these anomalies is critical for safety and plays a vital role in the successful and effective completion of the mission. However, identifying the moment where there is an anomaly or the direct cause of an anomaly is challenging.

A robust anomaly detection system for UAVs requires a holistic approach, evaluating sensor inputs collectively rather than individually. The detection mechanism should be capable of identifying three primary types of anomalies: outlier, contextual, and collective. Outlier anomalies, also known as point anomalies, are individual data points that deviate significantly from other observations or from the expected norm \cite{sun2017novel}. For example, a single, isolated spike in a magnetometer reading caused by temporary electromagnetic interference would be an outlier anomaly. A contextual anomaly occurs when a data instance is invalid within its specific operational context, even though the value itself may appear valid in isolation \cite{alos2021using}. An example is the mean acceleration on the z-axis deviating significantly from the gravitational constant while the UAV is in stable flight, as its orientation should keep the z-axis perpendicular to the ground. Collective anomalies involve a set of data points that are individually normal but become anomalous when analyzed together \cite{khalastchi2011online}. For instance, a notable increase in altitude while the thrust is simultaneously decreasing would constitute a collective anomaly, as these two actions are physically contradictory under normal flight conditions. 

Considering all the possible anomaly reasons, both the EKF outcomes and input sources must be monitored to detect and classify any anomalies. Developing a monitoring system requires prior data for analysis. This paper addresses this gap and presents a dataset to facilitate further investigation within state estimation anomaly research, along with introducing the classification of state estimation anomalies in UAVs.

\usetikzlibrary{shapes.geometric, arrows.meta, positioning, calc}

\definecolor{sensorblue}{RGB}{235, 245, 255}
\definecolor{ekfpurple}{RGB}{243, 230, 255}
\definecolor{stategreen}{RGB}{230, 255, 240}
\definecolor{borderblue}{RGB}{0, 102, 204}
\definecolor{borderpurple}{RGB}{102, 0, 204}
\definecolor{bordergreen}{RGB}{0, 153, 76}

\begin{figure}
    \centering

    \begin{adjustbox}{width=\linewidth}
    \includegraphics{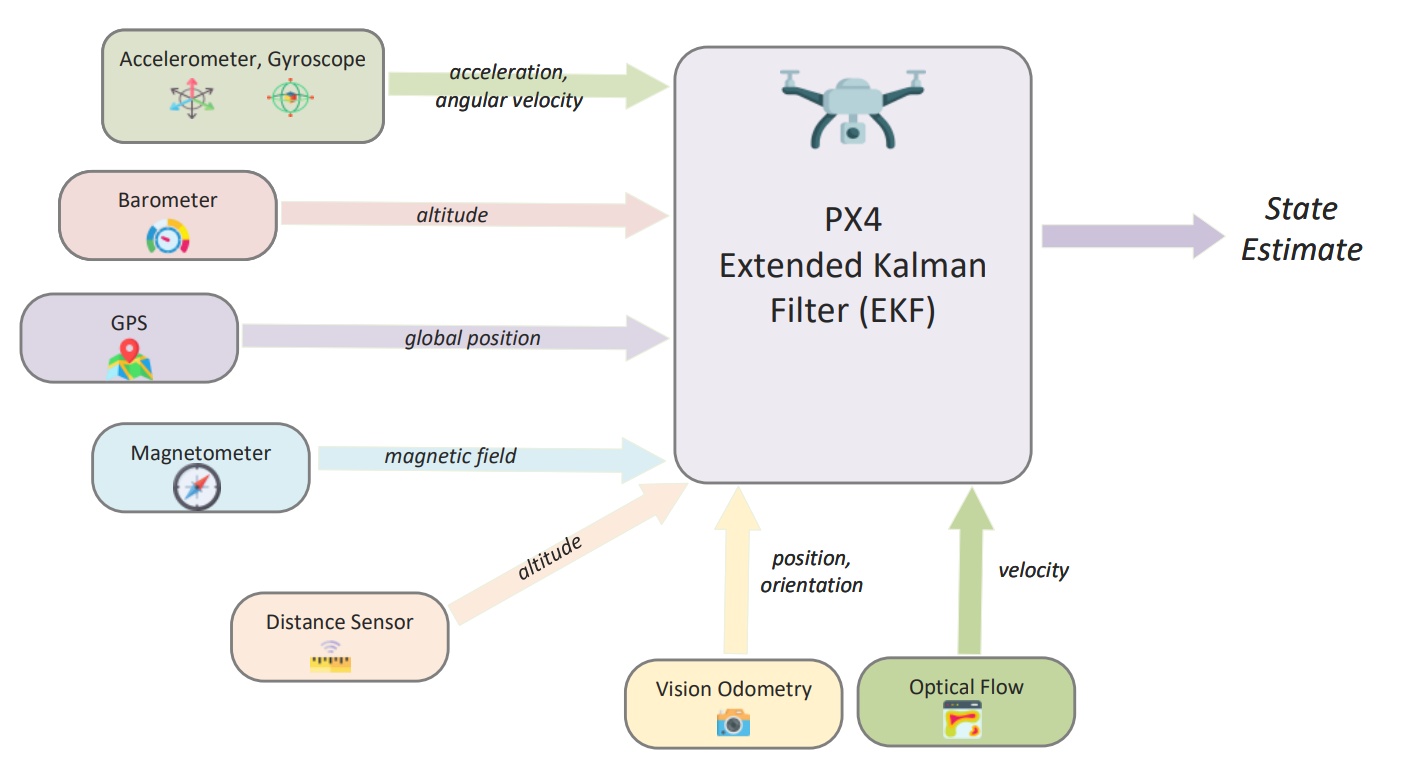}
    \end{adjustbox}
    \caption{PX4 EKF Structure }
    \label{px4_ekf}
\end{figure}

\section{UAV Models and Working Environment} \label{working_environment}
This section details the data collection methodology and the working environment, including the hardware components used. The UAV-SEAD dataset consists of real-world flight data and provides detailed flight logs. This data was collected not only in controlled laboratory conditions but also in a variety of real field environments such as warehouses, indoor sports halls, hangars, and outdoor spaces.

    \subsection{Hardware and Autopilot} \label{hardware_autopilot}

    The open source autopilot platform PX4 was employed during the data collection phase. PX4, in particular, has gained widespread adoption in both academic and industrial communities due to its active development and capabilities. The autopilot system is compatible with dedicated hardware platforms such as Pixhawk or can be integrated into custom embedded systems. In this study, Pixhawk 2 Cube and Pixhawk 4 Mini modules were selected with the PX4 autopilot firmware installed.

\begin{figure}[h]
    \centering
    \begin{subfigure}{0.18\textwidth}
        \centering
        \includegraphics[width=\textwidth,height=2.5cm]{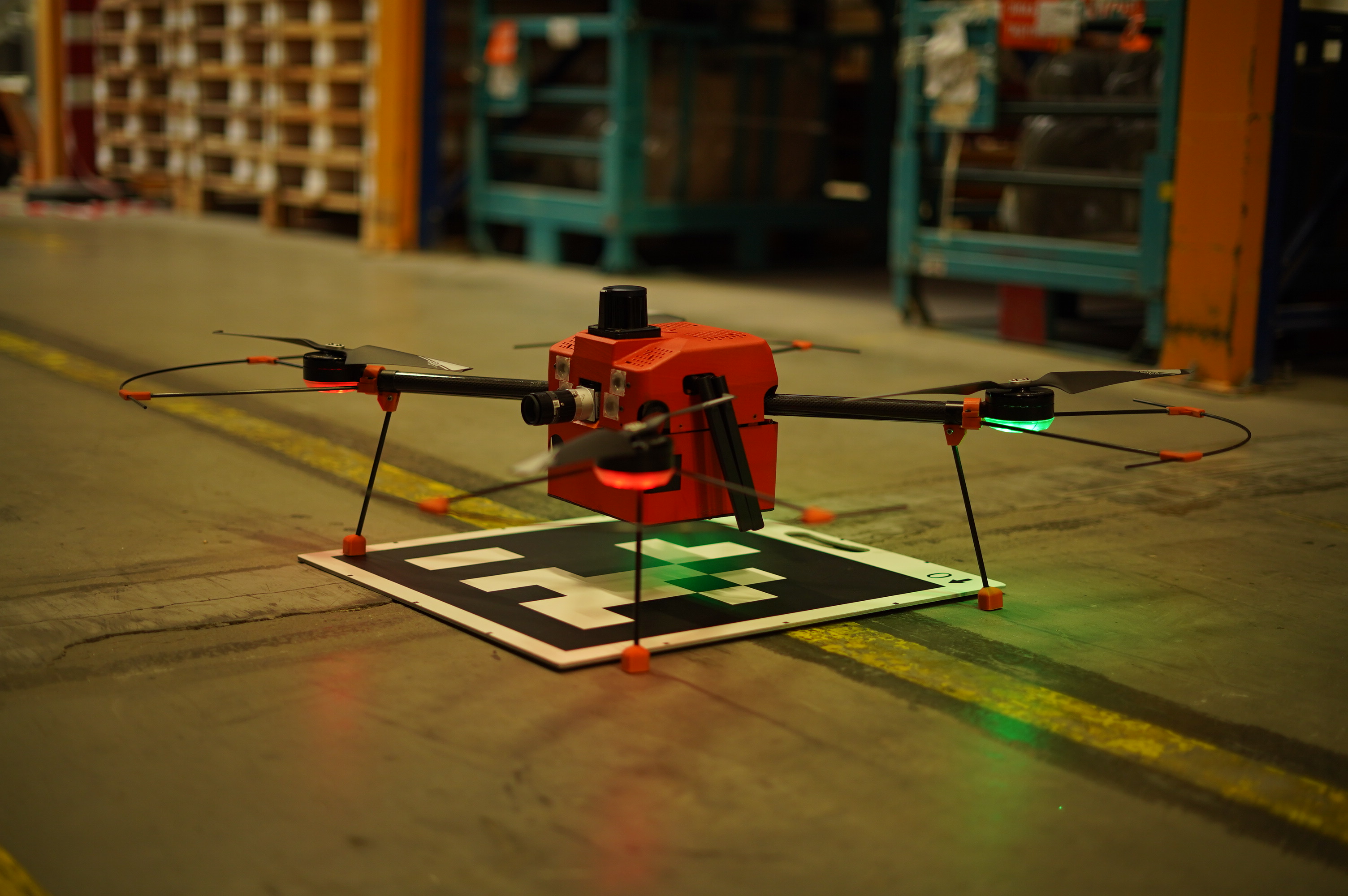}
    \end{subfigure}
    \begin{subfigure}{0.18\textwidth}
        \centering
        \includegraphics[width=\textwidth,height=2.5cm]{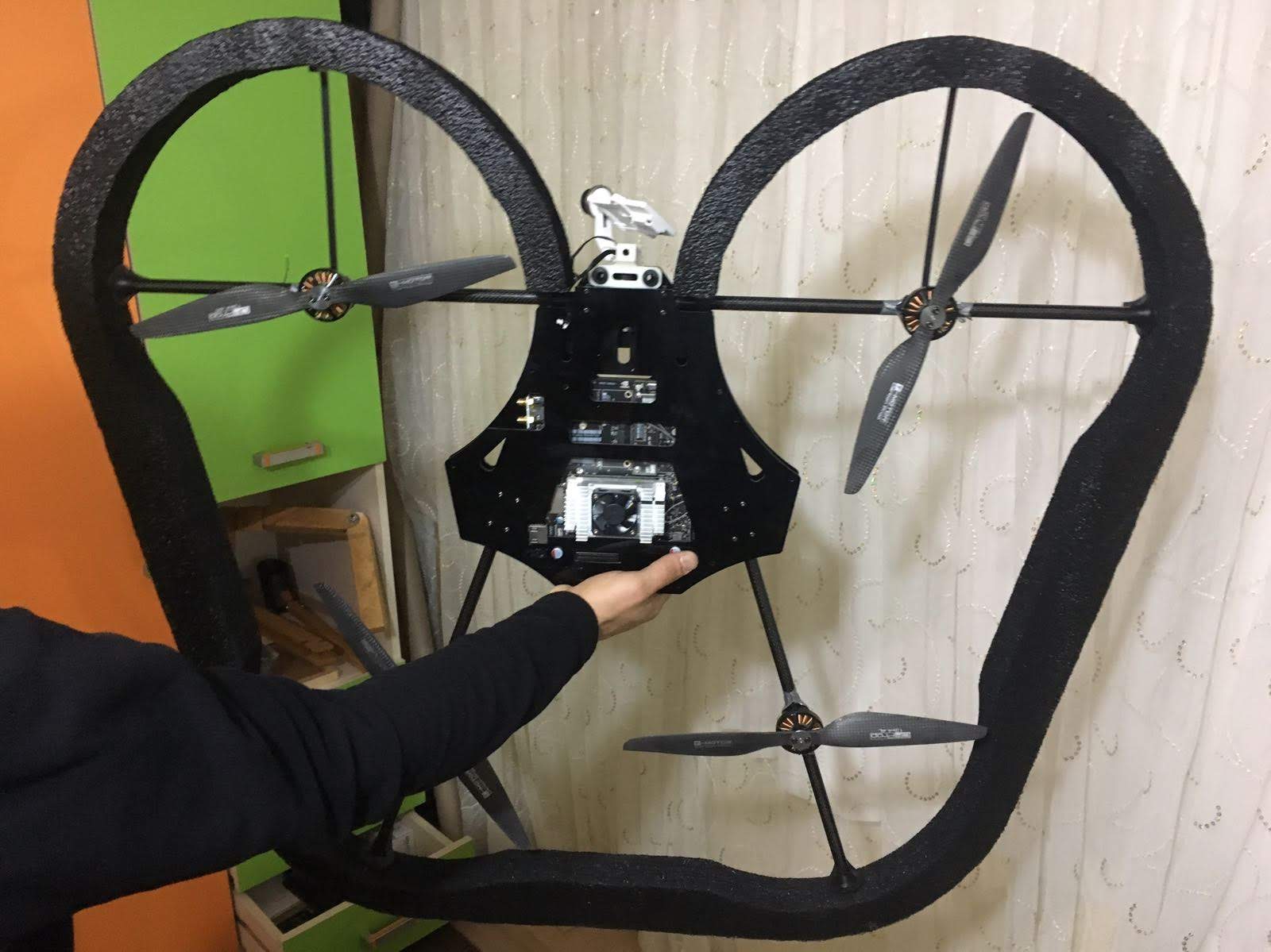}
    \end{subfigure}
    \begin{subfigure}{0.18\textwidth}
        \centering
        \includegraphics[width=\textwidth,height=2.5cm]{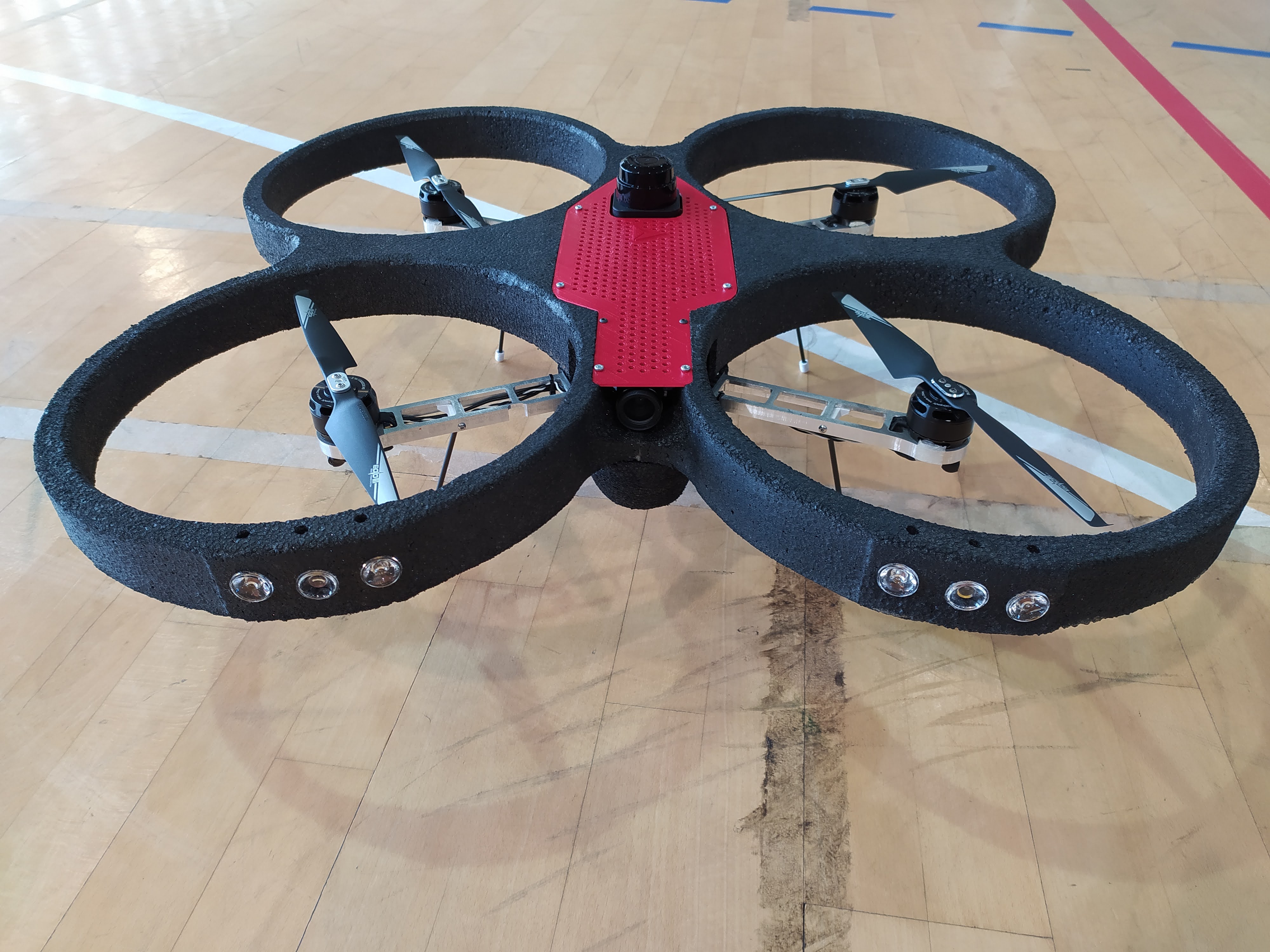}
    \end{subfigure}
    \begin{subfigure}{0.18\textwidth}
        \centering
        \includegraphics[width=\textwidth,height=2.5cm]{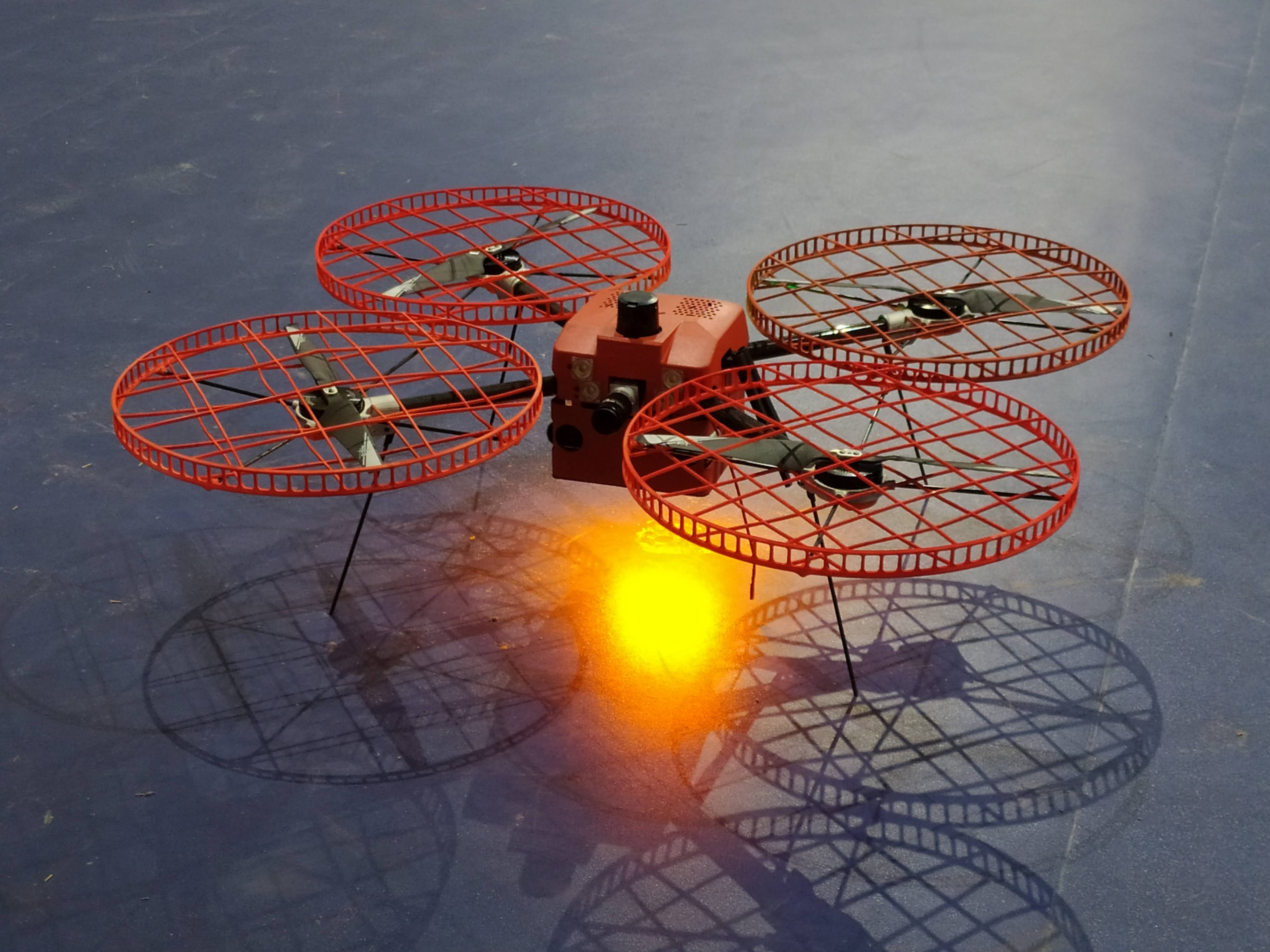}
    \end{subfigure}

    \vspace{1mm} 

    \begin{subfigure}{0.24\textwidth}
        \centering
        \includegraphics[width=\textwidth,keepaspectratio]{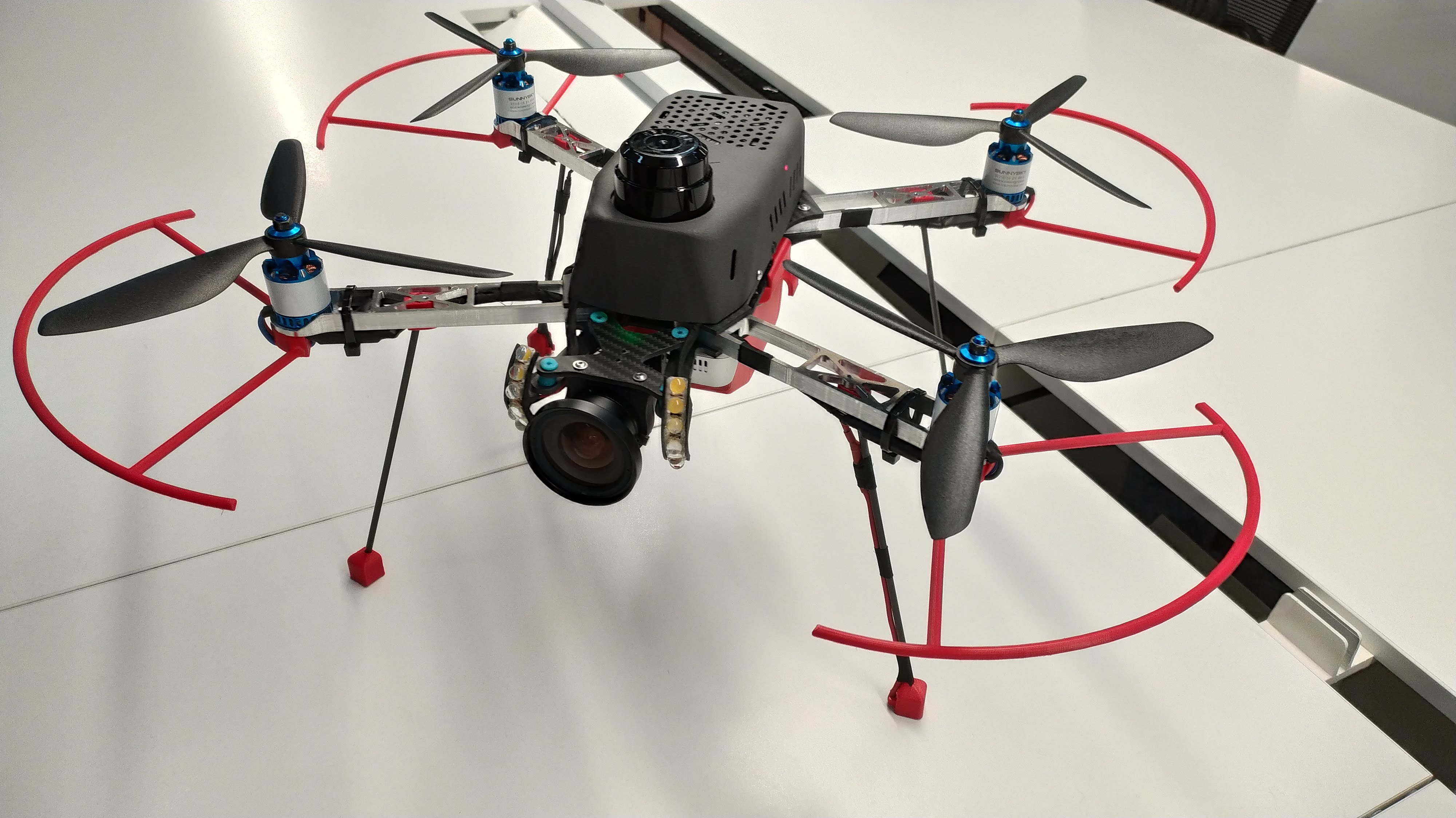}
    \end{subfigure}

    \caption{UAV Models}
    \label{uav_models}
\end{figure}

    In this study, six distinct configurations of the quadrotor airframes, each equipped with a different combination of onboard sensors, are used to capture a diverse set of flight behaviors aimed at identifying naturally occurring anomalies (see Figure~\ref{uav_models}). Each UAV airframe in this study is a custom design, developed by researchers, incorporating distinct sensor suites to explore a variety of state estimation scenarios. These frames are specifically engineered for indoor applications and feature various propeller cover configurations to evaluate diverse airflow and stability conditions. Variations in vehicle weight, payload capacity, and propeller dimensions ensure structural diversity across the dataset, enhancing its generalizability for different operational environments.

    PX4 supports a variety of input sources, including GPS, triple redundant IMUs (comprising accelerometers and gyroscopes), an internal magnetometer, barometer, vision odometry sources, optical flow sensors, and distance sensors, particularly for multirotor platforms. While most sensors can be directly connected to the Pixhawk, external position sources like vision odometry and optical flow must be computed on a companion computer or specialized hardware capable of generating position estimates. These outputs are transmitted to PX4 via interfaces such as UART or telemetry modules. Companion computers can run Simultaneous Localization and Mapping (SLAM) algorithms to generate vision odometry data by integrating sensors such as RGB cameras, depth cameras, stereo vision systems, or LIDAR. All aforementioned input sources are collectively considered for state estimation anomaly detection, and they are illustrated in Figure \ref{px4_ekf}.
    PX4 employs the EKF as its primary estimator for both multirotor and fixed-wing aerial platforms, fusing all incoming sensor modalities. This sensor fusion algorithm is implemented within the PX4 Estimation and Control Library (PX4-ECL) and is responsible for integrating data from multiple sensors to estimate the state of the vehicle, as shown in Figure \ref{px4_ekf}. 

    The data sources, their scopes, and the integrated hardware models of the UAV platforms used in the experiments are listed in Table \ref{table_sensors}.

\begin{table}[!t]
\caption{Sensor Specifications and Hardware Models}
\label{table_sensors}
\centering
\begin{tabular}{|>{\raggedright\arraybackslash}m{1.2cm}|>{\raggedright\arraybackslash}m{2.5cm}|>{\raggedright\arraybackslash}m{3.5cm}|}
    \hline
        \textbf{Sensor Type} & \textbf{Function} & \textbf{Integrated Hardware} \\
    \hline
        GPS & Provides global position, velocity, and time reference & HEX/ProfiCNC Here2, DroneCAN M8N \\
    \hline
        IMU & Measures linear acceleration and angular velocity & MPU9250, ICM20608, ICM20948, ICM20648 \\
    \hline
        Barometer & Measures altitude based on atmospheric pressure & MS5611 \\
    \hline
        Optical Flow & Provides velocity estimates via relative movement. & PX4FLOW \\
    \hline
        Distance Sensor & Used for height estimation & Garmin LidarLite v3, Benewake TFMini/Plus, TF02-Pro \\
    \hline
        External Position & Position estimates computed externally & Slamtec RPLidar S1, Hokuyo UST20LX, Intel Realsense D435i/T265, ZED, MarvelMind \\
    \hline
    \end{tabular}
\end{table}
    
    The EKF processes incoming sensor data by predicting the system’s state evolution and subsequently correcting the estimate based on new measurements. PX4-ECL allows users to configure input delays for each sensor through parameter settings, enabling synchronization with the estimator. Additionally, multiple sensor sources —such as GPS, barometer, distance sensor, and external position source— can be selectively enabled for height estimation. 

    The EKF implementation in the PX4 autopilot system requires data from an Inertial Measurement Unit (IMU) to estimate the vehicle's position; other sensor inputs are optional and can be configured based on the vehicle design and operational requirements. Certain input sources can be selectively disabled under specific operational conditions. For example, a distance sensor used to measure height above ground level may be deactivated after the vehicle exceeds a predefined altitude, also visual odometry input may be disabled during failsafe events to ensure system stability.

    Sensor inputs are susceptible to noise, environmental disturbances, or hardware faults, which may result in abnormal or erroneous data being propagated to the EKF output. Detecting such anomalies by analyzing individual sensor streams in isolation is often impractical, as faulty inputs can still appear nominal while being inconsistent with the true state of the system. Therefore, it is essential to evaluate the full set of input sources collectively and perform a comprehensive multimodal analysis of the estimated state.

    \subsection{Data Collection and Annotations}

    This study presents a dataset that incorporates various anomalous sensor inputs. This dataset is designed primarily for the purpose of detecting anomalies in the state estimation. The flight logs were not subjected to any artificial anomaly injection, preserving the reflection of real circumstances and actual behavior of UAVs in the recorded data and enhancing their suitability for a wide range of research applications. The logs were collected throughout the development and operational phases of the UAVs, reflecting a spectrum of maturity levels of the system and capturing a variety of real-world anomalies. Given the diversity of UAV models and the heterogeneity of sensor configurations presented in Section \ref{hardware_autopilot}, the dataset represents generalized UAV flight behaviors across multiple scenarios. As a result, this work provides the literature with a comprehensive and realistic dataset depicting both normal and anomalous flight conditions to support the development and evaluation of state estimation anomaly detection systems.
        
    The UAV-SEAD comprises a total of 1396 flights, amounting to over 52 hours of flight time. A summary of the flight duration statistics is presented in Table \ref{dataset_statistics}. Out of the total flight time, approximately 51 hours and 30 minutes include external position data, while global position data are available for 4 hours and 35 minutes. Some flight records contain only global position data, whereas some incorporate both external and global positioning sources. A similar variation is observed in the usage of distance sensors and barometers for altitude estimation. UAV-SEAD is publicly available and hosted on the Hugging Face Hub (\cite{uav_sead_dataset}).
    
    The dataset annotations were created using a custom-designed annotation toolbox and can be publicly accessible \cite{aykut_kabaoglu_2025_17220629}. The open-source annotation package enables users to create timestamp-based annotations. The data can be processed using downsampling or upsampling techniques to create uniform-sized data frames. The annotation toolbox enables customizable plotting of the selected data topics to assist in the labeling process. Users can select and visualize combinations of topics based on their annotation objectives. By default, the toolbox provides visualizations for each axis of the vehicle’s position: X, Y, and Z. For instance, the X-axis position plot includes estimated local position, visual odometry, and GPS-derived position. The Z-axis position plot incorporates estimated local position, distance sensor readings, visual odometry, barometric altitude, and GPS altitude. Visual odometry and GPS data for all three axes are also plotted separately. Additional figures present other flight parameters, such as manual control inputs, autopilot actuator outputs, motor pulse-width modulation (PWM) signals, etc. 

    UAV-SEAD offers several notable advantages, including a large volume of real-world flight data, the use of diverse sensor modalities, testing across various airframe configurations, exposure to naturally occurring anomalies (without artificial injection), representation of a wide range of anomaly categories, and the inclusion of software-related anomalies. Note that the UAV-SEAD includes a relatively small number of outdoor flights with occasional anomalies in global positioning data because working areas are mostly indoor. However, the dataset can be expanded and enriched by incorporating additional flight data from the community easily with the use of the annotation package. The primary aim of this dataset is to establish a foundational, large-scale benchmark for state estimation anomaly research; while the provided labels are not intended to define exact anomaly onset or offset times, they delineate anomalous regions of interest, serving to guide research focus, foster community engagement, and increase awareness of state estimation anomalies in UAV systems.

    \begin{table}
    \caption{Flight Duration Overview \label{tbl:flight_duration}}
    \label{dataset_statistics}
    \centering
    \resizebox{0.85\columnwidth}{!}{
        \begin{tabular}{|l|c|}
        \hline
        \textbf{Type} & \textbf{Time (hh:mm:ss)} \\ \hline
        Total flight duration         & 52:20:05 \\
        External position data duration    & 51:28:34 \\
        Global position data duration      & 04:35:24 \\
        Distance sensor data duration      & 51:35:05 \\
        Barometer altitude data duration   & 26:23:45 \\ \hline
        \end{tabular}
        }
    \end{table}
    \vspace{1em}


\begin{figure*}[htbp]
    \centering
    \begin{tikzpicture}[
      scale=1.25,
      transform shape,
      level 1/.style={
        sibling distance=30mm, 
        level distance=10mm,
        every node/.append style={fill=blue!20}
      },
      level 2/.style={
        sibling distance=0mm, 
        level distance=10mm,
        every node/.append style={fill=green!10}
      },
      level 3/.style={
        sibling distance=0mm, 
        level distance=10mm,
        every node/.append style={}
      },
      every node/.style={
        draw, 
        rounded corners, 
        font=\tiny
      },
      edge from parent/.style={
        draw, 
        -{Latex[length=1mm]}, 
        behind path
      }
    ]
     \node [fill=red!20] (root) {State Estimation Anomalies}
        child[level distance=10mm] {node {Mechanical and Electrical Anomalies}
            child[edge from parent path={(\tikzparentnode.south) -- ++(-0.5,0) |- (\tikzchildnode.west)}, level distance=5mm, anchor=west] {node {Acceleration}}
            child[edge from parent path={(\tikzparentnode.south) -- ++(-0.5,0) |- (\tikzchildnode.west)}, level distance=10mm, anchor=west] {node {Angular Velocity}}
            child[edge from parent path={(\tikzparentnode.south) -- ++(-0.5,0) |- (\tikzchildnode.west)}, level distance=15mm, anchor=west] {node {Voltage}}
            child[edge from parent path={(\tikzparentnode.south) -- ++(-0.5,0) |- (\tikzchildnode.west)}, level distance=20mm, anchor=west] {node {Current}}
            child[edge from parent path={(\tikzparentnode.south) -- ++(-0.5,0) |- (\tikzchildnode.west)}, level distance=25mm, anchor=west] {node [fill=green!70] {Motor Outputs}}
            child[edge from parent path={(\tikzparentnode.south) -- ++(-0.5,0) |- (\tikzchildnode.west)}, level distance=30mm, anchor=west] {node [fill=green!70] {Actuator Controls}}
            child[edge from parent path={(\tikzparentnode.south) -- ++(-0.5,0) |- (\tikzchildnode.west)}, level distance=35mm, anchor=west] {node [fill=green!70] {Attitude Rates}}
        }
        child[level distance=10mm] {node {External Position Anomalies}
            child[edge from parent path={(\tikzparentnode.south) -- ++(-0.5,0) |- (\tikzchildnode.west)}, level distance=5mm, anchor=west] {node {Visual Odometry}}
            child[edge from parent path={(\tikzparentnode.south) -- ++(-0.5,0) |- (\tikzchildnode.west)}, level distance=10mm, anchor=west] {node {Optical Flow}}
            child[edge from parent path={(\tikzparentnode.south) -- ++(-0.5,0) |- (\tikzchildnode.west)}, level distance=15mm, anchor=west] {node {GPS Position}}
            child[edge from parent path={(\tikzparentnode.south) -- ++(-0.5,0) |- (\tikzchildnode.west)}, level distance=20mm, anchor=west] {node {Velocity}}
            child[edge from parent path={(\tikzparentnode.south) -- ++(-0.5,0) |- (\tikzchildnode.west)}, level distance=25mm, anchor=west] {node [fill=green!70] {Actuator Controls}}
            child[edge from parent path={(\tikzparentnode.south) -- ++(-0.5,0) |- (\tikzchildnode.west)}, level distance=30mm, anchor=west] {node [fill=green!70] {EKF Estimated Position}}
        }
        child[level distance=10mm] {node {Global Position Anomalies}
            child[edge from parent path={(\tikzparentnode.south) -- ++(-0.5,0) |- (\tikzchildnode.west)}, level distance=5mm, anchor=west] {node {GPS Position}}
            child[edge from parent path={(\tikzparentnode.south) -- ++(-0.5,0) |- (\tikzchildnode.west)}, level distance=10mm, anchor=west] {node [fill=green!70] {Actuator Controls}}
            child[edge from parent path={(\tikzparentnode.south) -- ++(-0.5,0) |- (\tikzchildnode.west)}, level distance=15mm, anchor=west] {node [fill=green!70] {EKF Estimated Position}}
        }
        child[level distance=10mm] {node {Altitude Anomalies}
            child[edge from parent path={(\tikzparentnode.south) -- ++(-0.5,0) |- (\tikzchildnode.west)}, level distance=5mm, anchor=west] {node {Pressure}}
            child[edge from parent path={(\tikzparentnode.south) -- ++(-0.5,0) |- (\tikzchildnode.west)}, level distance=10mm, anchor=west] {node {Terrain Distance}}
            child[edge from parent path={(\tikzparentnode.south) -- ++(-0.5,0) |- (\tikzchildnode.west)}, level distance=15mm, anchor=west] {node {GPS Altitude}}
            child[edge from parent path={(\tikzparentnode.south) -- ++(-0.5,0) |- (\tikzchildnode.west)}, level distance=20mm, anchor=west] {node {Visual Odometry}}
            child[edge from parent path={(\tikzparentnode.south) -- ++(-0.5,0) |- (\tikzchildnode.west)}, level distance=25mm, anchor=west] {node [fill=green!70] {EKF Estimated Position}}
            child[edge from parent path={(\tikzparentnode.south) -- ++(-0.5,0) |- (\tikzchildnode.west)}, level distance=30mm, anchor=west] {node [fill=green!70] {Thrust}}
        }
        ;
    \end{tikzpicture}
    \caption{State estimation anomalies and their associated observation sources.} {\textit{Purple: Anomaly class, Light Green: Sensor observation, Dark Green: Autopilot output.}}
    \label{anomaly_categories_inputs}
\end{figure*}

\section{Classification of State Estimation Anomalies} \label{anomaly_classification}

In addition to anomaly detection, classification of the anomaly is essential for evaluating UAV system behavior and ensuring successful flight operations.
State estimation anomalies can arise from a wide range of sources. We propose a new structured classification of state estimation anomalies, organized into four broad classes: Mechanical and Electrical Anomalies, External Position Anomalies, Global Position Anomalies, and Altitude Anomalies to distinguish among their sources, support targeted diagnostics, and facilitate the development of more robust detection and response mechanisms in UAV systems.
These anomalies can be detected using various sensory data streams. Anomalies may originate from a single sensor or from interactions among multiple sensor outputs. Moreover, a single sensor can contribute to the occurrence of multiple anomalies. The introduced classification framework for state estimation anomalies is agnostic to the hardware used, and can be generalized across various hardware, enabling broader applicability in anomaly detection research.

Our reference platform Pixhawk/PX4 provides data on acceleration, linear speed, angular speed, visual odometry, optical flow, GPS position, altitude, distance sensor values, magnetic field, voltage/current, actuator outputs, and fused state estimation output (local position), all of which can be collected through autopilot logs.

The state estimation anomaly classification and observation source tree is presented in Figure \ref{anomaly_categories_inputs}, comprising four primary categories. Each anomaly class is associated with specific input sources and major sources for that class are represented as branches of the tree. Light green nodes in branches represent sensor observations, and dark green nodes show autopilot outputs. For instance, external position anomalies can be identified by cross-validating visual odometry values, GPS data, linear velocity measurements, actuator control outputs, and the EKF-fused position estimation. This structured representation aids in systematically associating sensor data and autopilot outputs with corresponding anomaly types, facilitating a comprehensive analysis of state estimation failures.

  \subsection{Mechanical and Electrical Anomalies}
Mechanical anomalies in UAVs can arise from mechanical instabilities, high vibration levels, improper sensor calibrations, and deviations in the center of mass.

Mechanical factors such as sensor misalignment, component displacement (e.g., loose screws or structural deformation), and variable propeller speeds due to mechanical wear or mass imbalance contribute to incorrect sensor readings and uneven force distribution. These issues adversely affect the stability and accuracy of the state estimation process. Moreover, instabilities in the attitude control loop can result in persistent oscillations or divergence in orientation.

Inertial Measurement Units (IMUs), which include gyroscopes and accelerometers, are particularly sensitive to vibration. Excessive vibrations introduce noise and bias into angular rate and linear acceleration measurements. Vibrations at certain frequencies can resonate with the UAV's frame or sensor, amplifying inaccuracies. This noise, when integrated over time, leads to significant inaccuracies in the estimation of velocity and position, especially affecting roll and pitch orientation.

Calibration errors in IMUs further exacerbate these problems. Such errors propagate through the integration process, contributing to drift in the estimated position. Over time, accumulated noise and sensor biases can lead to substantial deviations from the ground-truth, undermining reliable navigation and control.

Incorrectly aligned frames of reference (e.g., body frame vs. inertial frame) cause errors in sensor fusion. When the center of mass shifts, the thrust required from each motor to maintain stability changes. This alters the force distribution across the UAV. The UAV may experience unexpected moments about its roll, pitch, or yaw axes. Accelerations become biased due to unintended rotational components caused by the offset in the center of mass. The accelerometer assumes that the UAV's center of mass aligns with its geometric center if the offset is not set correctly. Deviations cause discrepancies between the measured and actual accelerations, leading to drift in velocity and position estimates. The flight controller compensates for the shifted center of mass by continuously adjusting motor outputs. This can introduce oscillations or unstable control behavior.

Electrical anomalies may occur due to excessive or unbalanced payload, actuator failures, and connected peripherals that fluctuate voltage, which may result in sudden voltage drops or current spikes. The battery may be unable to compensate for these changes, or the battery discharge rates may not be enough to fulfill the energy demand. 

Excessive payloads may result in increased energy demands. This leads to sudden voltage drops during high-thrust maneuvers. When the payload is positioned off-center, one side of the vehicle may require more thrust to stabilize. This uneven demand increases the current draw on ESCs (Electronic Speed Controllers).

A partial or total actuator failure may result in an electrical anomaly. A partially failing motor with increased resistance may draw inconsistent current. The battery may struggle to compensate for the fluctuating demand, leading to observable oscillations in the system state. 

A UAV equipped with a battery that has an insufficient discharge rate may be unable to handle the peak current demands during aggressive maneuvers, voltage drops can shut down critical components such as the flight controller or GPS module, or cause them to restart mid-flight, potentially leading to unexpected position shifts.

Mechanical and electrical anomalies can be identified by analyzing data from accelerometers, gyroscopes, voltage / current measurements, motor outputs, actuator controls, and attitude(roll, pitch, yaw) rates. Due to the similarity in behavior and observation sources of mechanical and electrical anomalies, they are grouped into a single anomaly class.
  
  \subsection{External Position Anomalies}  
External position refers to UAV position information obtained from sensors, systems, or data sources external to the vehicle rather than from estimates generated internally through onboard sensor fusion. The PX4 system can receive external position data from sources such as visual odometry or optical flow. PX4 can be fed with position data using the visual odometry topic. Visual odometry data can be generated using visual SLAM, UWB(Ultra Wide Band) positioning systems, LIDAR SLAM, etc. PX4 uses custom hardware for optical flow that can be directly connected to the Pixhawk, and these devices can directly feed PX4 with flow data as an external position source.

SLAM systems are among the most widely used methods for generating external position data, and anomalies can be caused by software errors in algorithms or fast manoeuvres where visual features cannot be traced properly. Both visual SLAM systems and optical flow systems rely on sufficient lighting for accurate tracking. Poor lighting conditions can result in inaccurate position estimates. Additionally, at high altitudes, cameras fail to perceive sufficient visual features and lose the tracking. Also, failures in sensors themselves may cause a complete loss of external position data.

As a result, these conditions result in abrupt and significant changes in the reported position and can disrupt the UAV's navigation or control. These anomalies are classified as external position anomalies, which can be identified by analyzing visual odometry data, optical flow, GPS position, velocity, actuator controls, and EKF estimated position.

  \subsection{Global Position Anomalies}
Global position anomalies in UAVs are often caused by GPS glitches, weak signal reception, or hardware failures. Additionally, spoofed or manipulated GPS data can be injected into the PX4 system through custom positioning hardware or a companion computer, providing false latitude, longitude, and altitude values to the autopilot.

Multipath effects, where GPS signals are reflected off buildings, terrain, or other surfaces, can introduce time delays and lead to inaccurate position estimates. Such distortions may lead to erratic flight behavior, loss of position lock, or incorrect coordinate reporting. Furthermore, radio frequency interference can disrupt the reception of valid GPS signals, while hardware issues—such as damaged antennas, degraded electronic components, or low-quality GPS receivers—can produce inconsistent or erroneous global position data.

Global position anomalies typically manifest as sudden jumps in the estimated position. Moreover, GPS updates are inherently discrete, and the time intervals between successive observations may be too long to support accurate and reliable state estimation, especially in high-dynamics scenarios. These anomalies can be identified by observing the GPS position, actuator controls, and EKF estimated position.

  \subsection{Altitude Anomalies}
  
Altitude anomalies in UAVs may arise from errors in pressure measurements, improper sensor calibration, or environmental interference—particularly when using distance sensors. These issues can significantly impair the accuracy of altitude estimation, especially when barometers or distance sensors are the primary sources for altitude estimation. 

Barometric altitude measurements are susceptible to disruption from sudden atmospheric pressure changes, which may occur due to rapid weather fluctuations or strong airflow disturbances caused by nearby UAVs or rotors. Distance sensors, such as LiDAR or ultrasonic devices, can also produce erroneous altitude readings when the UAV is operating above highly reflective surfaces, such as water or glass. These surfaces may scatter or distort emitted signals, leading to unreliable measurements. Furthermore, miscalibrated barometers or distance sensors can introduce long-term drift in altitude estimates, potentially causing the UAV to maintain incorrect flight altitudes.

Although distance sensors are commonly used for altitude estimation, their reliability diminishes at significant heights above ground due to inherent range limitations and decreased measurement accuracy. When the UAV exceeds the effective sensing range, the data provided by these sensors may become unreliable or unavailable, potentially leading to inaccurate altitude estimates.


The given examples can lead to a system-level state estimation anomaly, and identifying the exact cause of these abnormal situations is challenging, because outlier readings from a single sensor do not necessarily indicate a state estimation anomaly, and such anomalies may not be observable from any single sensor alone. Therefore, all sources, which are pressure, terrain distance, GPS altitude, visual odometry and thrust are considered together to detect collective and contextual abnormal behaviors.

\begin{table*}[h!]
    \caption{Flight statistics of class-based anomaly annotation.  \label{tbl:labeled_annotated_stats}Ratio 1 indicates the flight time labeled with the corresponding anomaly class relative to the cumulative flight time; Ratio 2 indicates the share of anomaly duration associated with a given class compared to the total annotated duration; Ratio 3 indicates the percentage of anomalous duration relative to the total flight time for that specific class.}
    \centering
    \resizebox{0.9\linewidth}{!}{
    \begin{tabular}{|l|c|cc|cc|c|}
    \hline
    \textbf{Class Labeling} & \makecell{\textbf{\# Flights} \\ (\# Logs)} & \makecell{\textbf{Total Flight Duration} \\ (hh:mm:ss)} & \textbf{Ratio 1} & \makecell{\textbf{Anomaly Duration} \\ (hh:mm:ss)} & \textbf{Ratio 2} & \textbf{Ratio 3} \\
    \hline
    Normal             & 900  & 38:24:19 & 71.91\% & --       & --  & -- \\
    Mechanical and Electrical & 47   & 02:12:38  & 4.14\%  & 00:12:43  & 8.36\% & 9.58\%  \\
    External Position  & 197  & 05:12:25  & 9.75\%  & 01:17:02  & 50.65\% & 24.65\% \\
    Global Position    & 41   & 01:40:42  & 3.14\%  & 00:36:23  & 23.92\% & 36.13\% \\
    Altitude           & 78   & 02:41:21  & 5.04\%  & 00:25:57  & 17.07\% & 16.08\% \\
    Uncategorized      & 141  & 03:13:09  & 6.03\%  & --       & --  & --   \\
    \hline
        \hline
    Cumulative Total & 1404 & 53:14:40 & 100\%     & 02:32:05  & 100\%  & 4.76\% \\
    \hline
\hline
    Actual Flight Statistics & 1396 & 52:20:05 & --      & --  & --  & 4.84\% \\
    \hline
    \end{tabular}
    }
\end{table*}

\section{UAV-SEAD Dataset} \label{uav_sead_dataset}
This section details the collected dataset, provides flight and classification statistics; additionally, describes how anomalous flights are annotated by giving the examples

The state estimation anomalies are identified by considering the divergence between the estimated state in the autopilot and the observed state in the real environment. Divergences such as drift, jitter, or sudden jumps are detectable through human observation, and the flights are annotated semantically. While visual tracking systems can provide ground truth in small-scale environments, obtaining precise ground-truth data through environmental tracking is infeasible in both outdoor and large-scale indoor environments. Since all sensory data sources are potential sources of anomalies, none can serve as absolute ground truth. Therefore, external human expert support is employed to label anomalous data by detecting divergences between the real and estimated states. Furthermore, data sources are cross-validated by investigating the logical relations and physical constraints between each topic described in Section \ref{working_environment}. The detailed evaluation process for these labels is illustrated through examples in Section \ref{example_anomalies}.

\subsection{Dataset Overview}

Table \ref{tbl:labeled_annotated_stats} summarizes the distribution and annotation statistics for class labels presented in Section \ref{anomaly_classification} of the UAV flight dataset. Out of 1396 flight logs, 900 (71.91\%) are categorized as normal flights corresponding to a total of 38 hours, 24 minutes, and 19 seconds of flight time. No detectable anomalies were identified in these flights during expert review. Anomalies are classified into four main classes: Mechanical and Electrical Anomaly, Altitude Anomaly, External Position Anomaly, and Global Position Anomaly. 

While UAV-SEAD provides a high-density focus on indoor operational environments, it also includes a targeted selection of outdoor flights to capture initial global positioning anomalies. This specialization makes the dataset an unparalleled resource for researchers focused on GPS-denied or constrained-space navigation. Furthermore, the dataset is designed as a scalable framework; through our provided open-source annotation package, the research community is invited to contribute additional flight data, ensuring that the repository continues to grow and diversify over time.

In the table, Ratio 1 indicates the corresponding flight duration for each class relative to the overall flight time. In other words, Ratio 1 represents the temporal footprint of class relative to the cumulative labeled duration (53:14:40), rather than the absolute flight time (52:20:05). This reflects the presence of overlapping anomalies where a single flight segment may contribute to multiple categories simultaneously. Ratio 2 represents the proportion of flight time labeled with a given anomaly class relative to the total anomalous flight time. Ratio 3 represents the proportion of total anomaly duration relative to the overall flight duration for the corresponding anomaly class. This metric highlights how frequently anomalous behavior occurs within flights labeled under a particular anomaly type.

The Mechanical and Electrical anomaly class combines faults related to onboard mechanical systems and power electronics. This class represents 4.14\% of the total flight duration, corresponding to 2 hours, 12 minutes, and 38 seconds, with 12 minutes and 43 seconds of those segments explicitly labeled as anomalous—representing 8.36\% of total annotated anomaly duration. This suggests that while these anomalies are relatively infrequent, their presence is notable.

External Position anomalies represent the most prominent class with its duration (1h 17m 02s), accounting for 50.65\% of all labeled anomaly segments. This indicates that these anomalies are both more persistent and more frequently encountered during flights, often manifesting in visual odometry or motion estimation systems.

Global Position anomalies are less frequent in terms of total flight time (3.14\%) but contribute 23.92\% to the total annotated anomaly duration. This disproportion suggests that GPS-related errors, while sparse, often involve prolonged or highly visible effects in state estimation.

Altitude anomalies account for 5.04\% of the total flight time and represent 17.07\% of the total annotated anomaly duration (25 minutes and 57 seconds). While Altitude Anomalies are more frequently observed in the dataset compared to Global Position Anomalies (78 vs. 41 flights), they account for significantly less footprint in total anomalous duration (25:57 vs. 36:23). This statistical difference suggests that Altitude Anomalies are generally more transient and may be recovered from more easily than Global Position Anomalies.

Uncategorized flights account for 141 records, representing 6.03\% of the total flight duration. These flights exhibit signs of abnormal behavior but could not be definitively classified under any of the predefined anomaly categories due to insufficient or ambiguous evidence. Despite the lack of categorical assignment, this subset is retained in the dataset as it may offer valuable insights to researchers. Including these records supports the goal of providing a comprehensive and broadly applicable dataset for the UAV research community.

It is important to note that a single flight may contain multiple anomalies occurring at different time intervals. As a result, overlaps exist in the reported statistics, particularly in the total flight durations and corresponding percentages for each anomaly class. Therefore, the total number of flights across classes in Table \ref{tbl:labeled_annotated_stats} exceeds 1396, since a single flight may be assigned multiple anomaly labels, just as the total labeled flights duration is less than the summation of the flight durations of each class. These overlapping flight statistics are represented under the cumulative total as 1404 flights with 53 hours, 14 minutes, and 40 seconds flight duration.

Altogether, 2 hours, 32 minutes, and 5 seconds of flight data are annotated as anomalous across the 1396 logs, forming the basis for a robust multiclass dataset. Segmentation and labeling were performed based on observable discrepancies in estimated versus expected motion, making this dataset a valuable resource for evaluating the robustness of UAV state estimation anomaly detection systems.

\subsection{Example Anomalies in the Dataset} \label{example_anomalies}
In this section, several example anomalies from the dataset are presented with their corresponding classes, together with the explanations of the labeling process. The anomaly durations are annotated in Figures \ref{fig:mechanical_combined}-\ref{fig:altitude_anomaly_combined} with red box timeframes. These annotations represent the same time interval for all figures and indicate an anomaly by investigating the relationships of the figures.

    \subsubsection{Mechanical and Electrical Anomalies}

\begin{figure*} 
        \centering
    
        \begin{subfigure}[b]{0.48\linewidth}
            \centering
            \includegraphics[width=\linewidth]{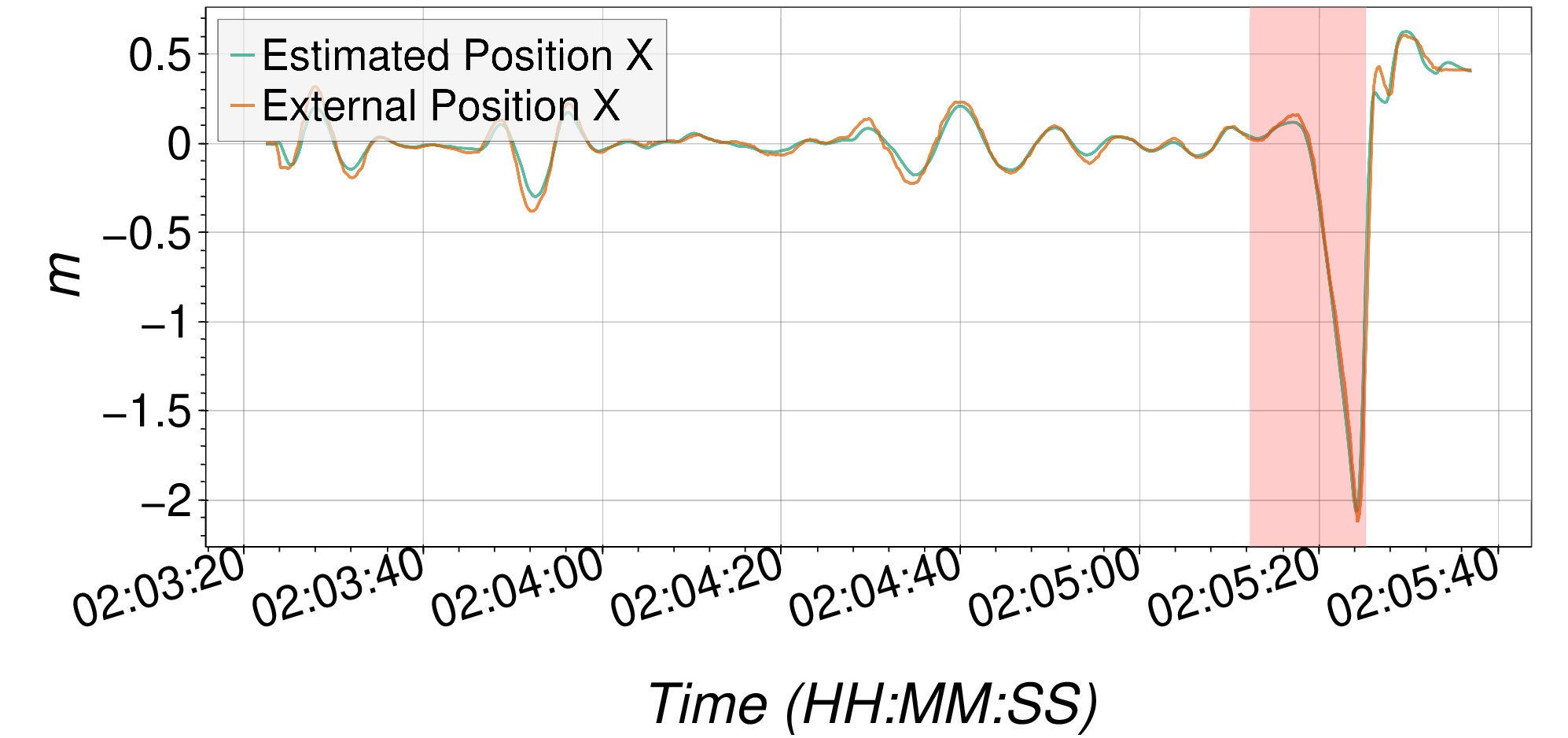}
            \caption{Position X}
            \label{mechanical_position_x}
        \end{subfigure}
        \hfill
        \begin{subfigure}[b]{0.48\linewidth}
            \centering
            \includegraphics[width=\linewidth]{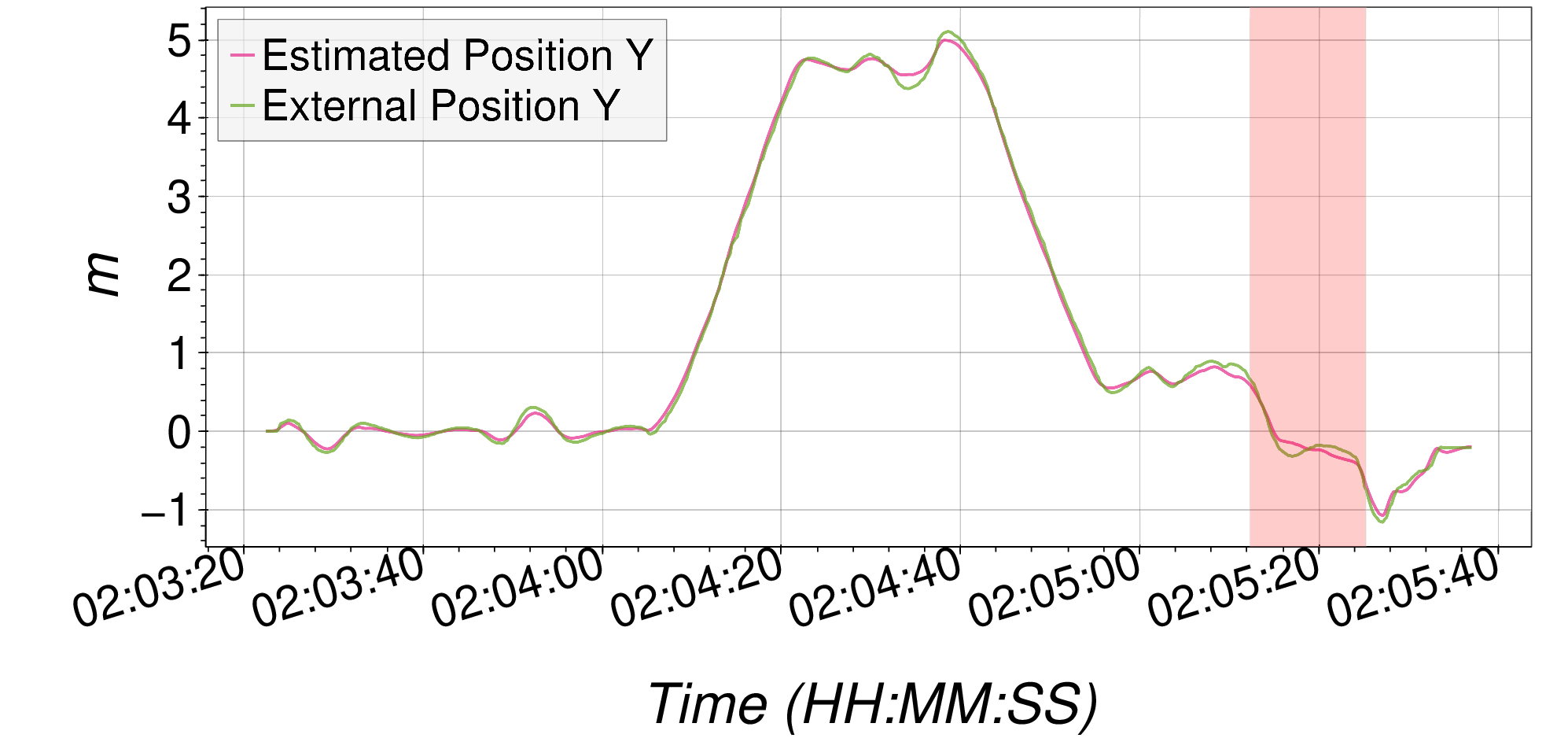}
            \caption{Position Y}
            \label{mechanical_position_y}
        \end{subfigure}
    
        \vspace{0.5em}

        \begin{subfigure}[b]{0.48\linewidth}
            \centering
            \includegraphics[width=\linewidth]{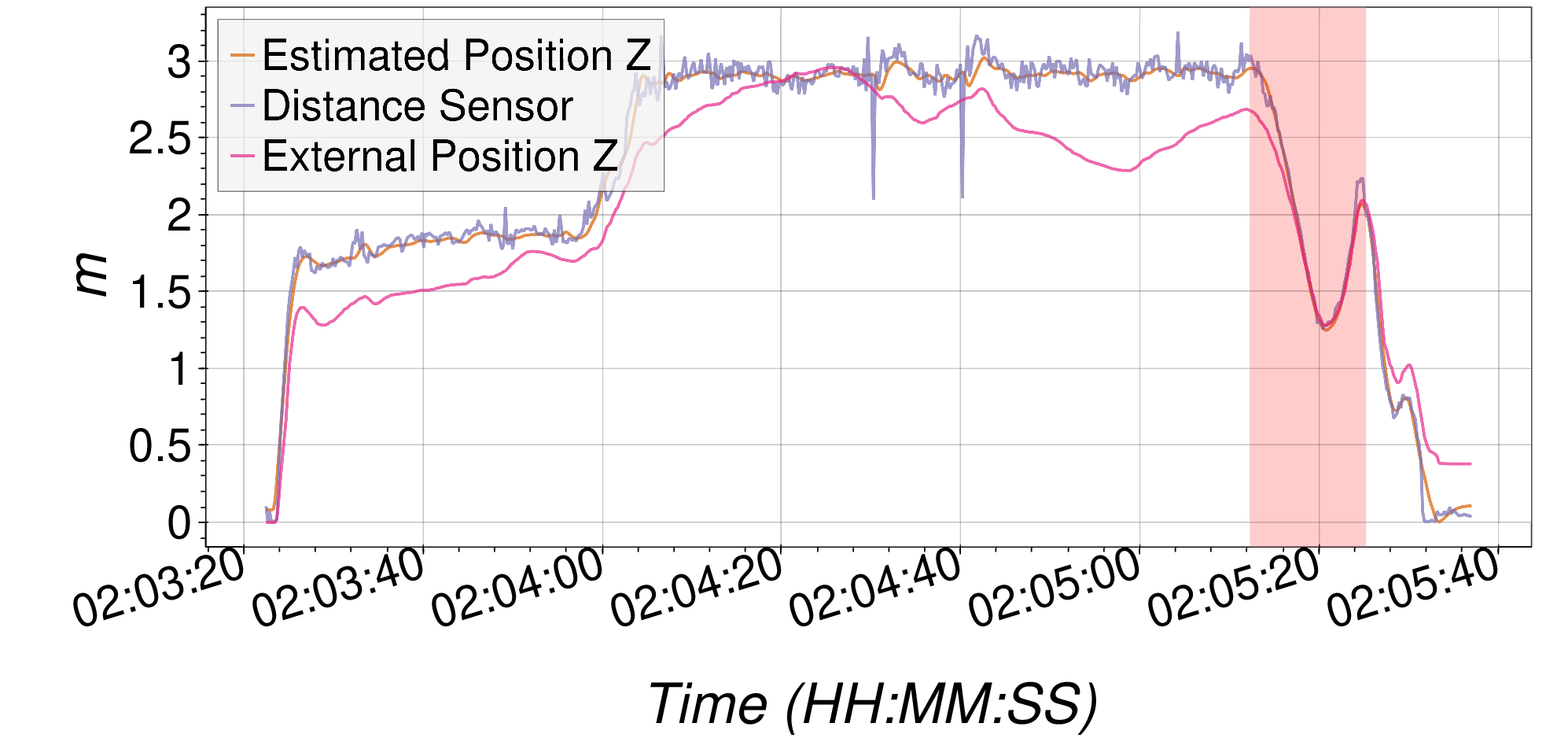}
            \caption{Position Z}
            \label{mechanical_position_z}
        \end{subfigure}
        \hfill
        \begin{subfigure}[b]{0.48\linewidth}
            \centering
            \includegraphics[width=\linewidth]{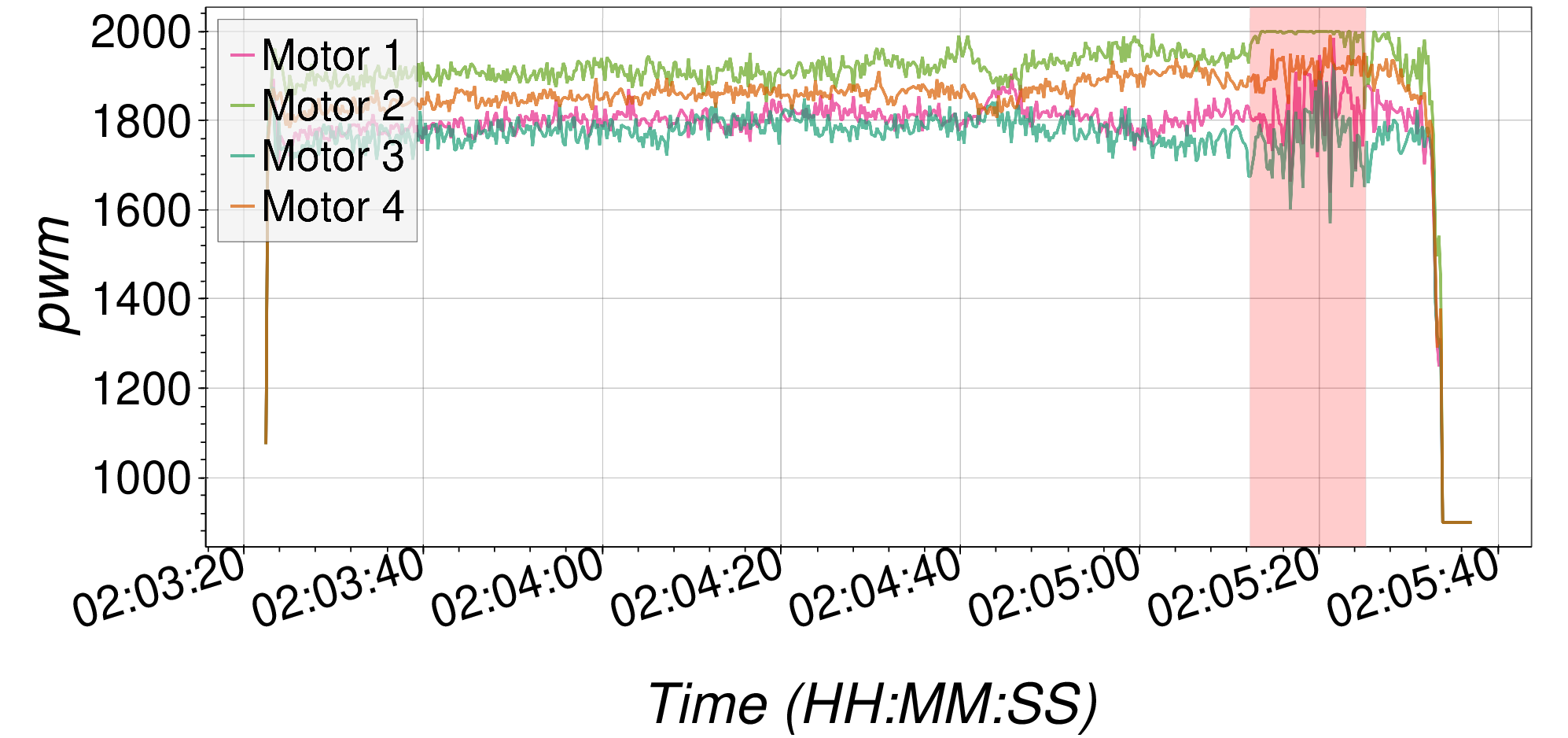}
            \caption{Motor Outputs}
            \label{mechanical_motor_outputs}
        \end{subfigure}

        \vspace{0.5em}        
        
        \begin{subfigure}[b]{0.48\linewidth}
            \centering
            \includegraphics[width=\linewidth]{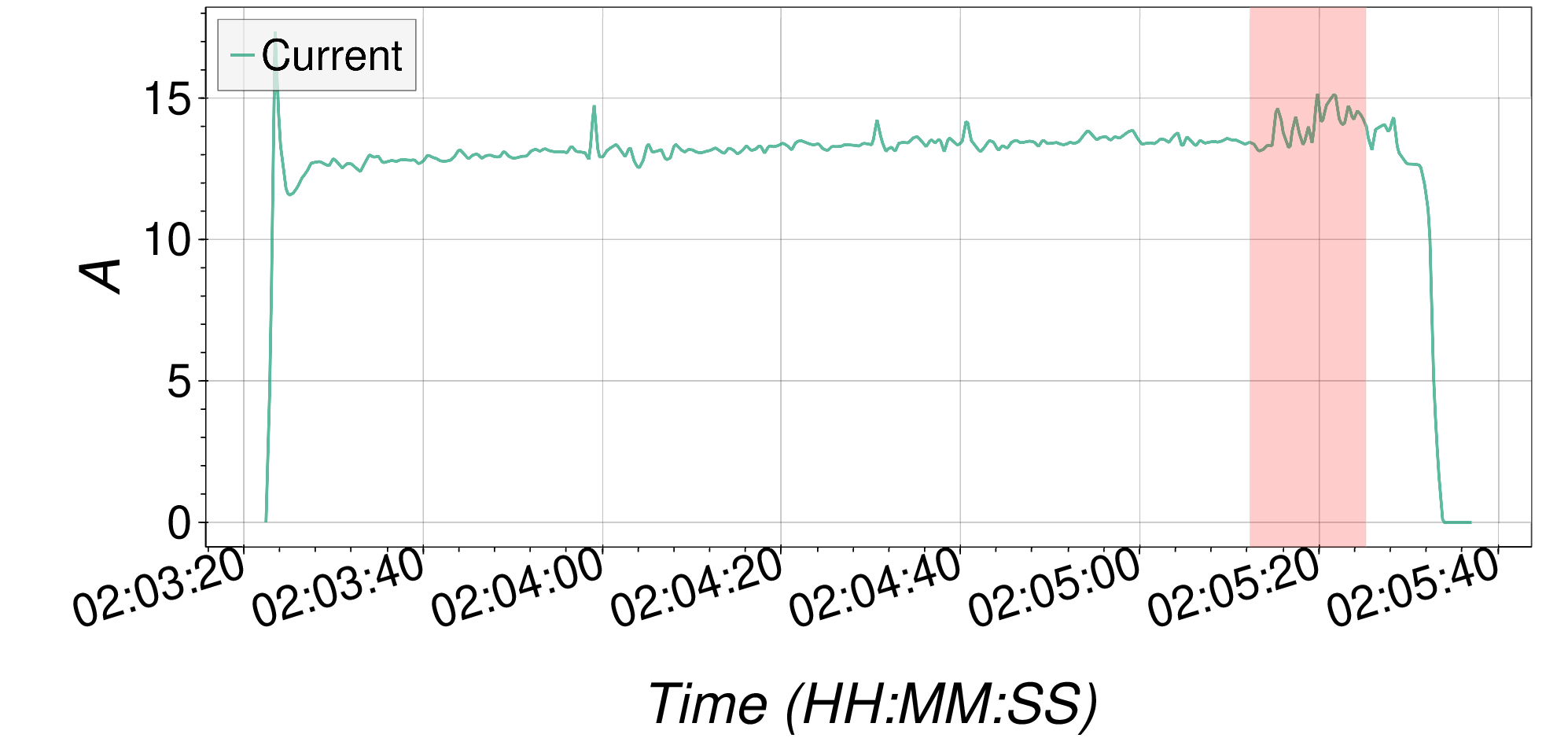}
            \caption{Current}
            \label{mechanical_current}
        \end{subfigure}
        \hfill        
        \begin{subfigure}[b]{0.48\linewidth}
            \centering
            \includegraphics[width=\linewidth]{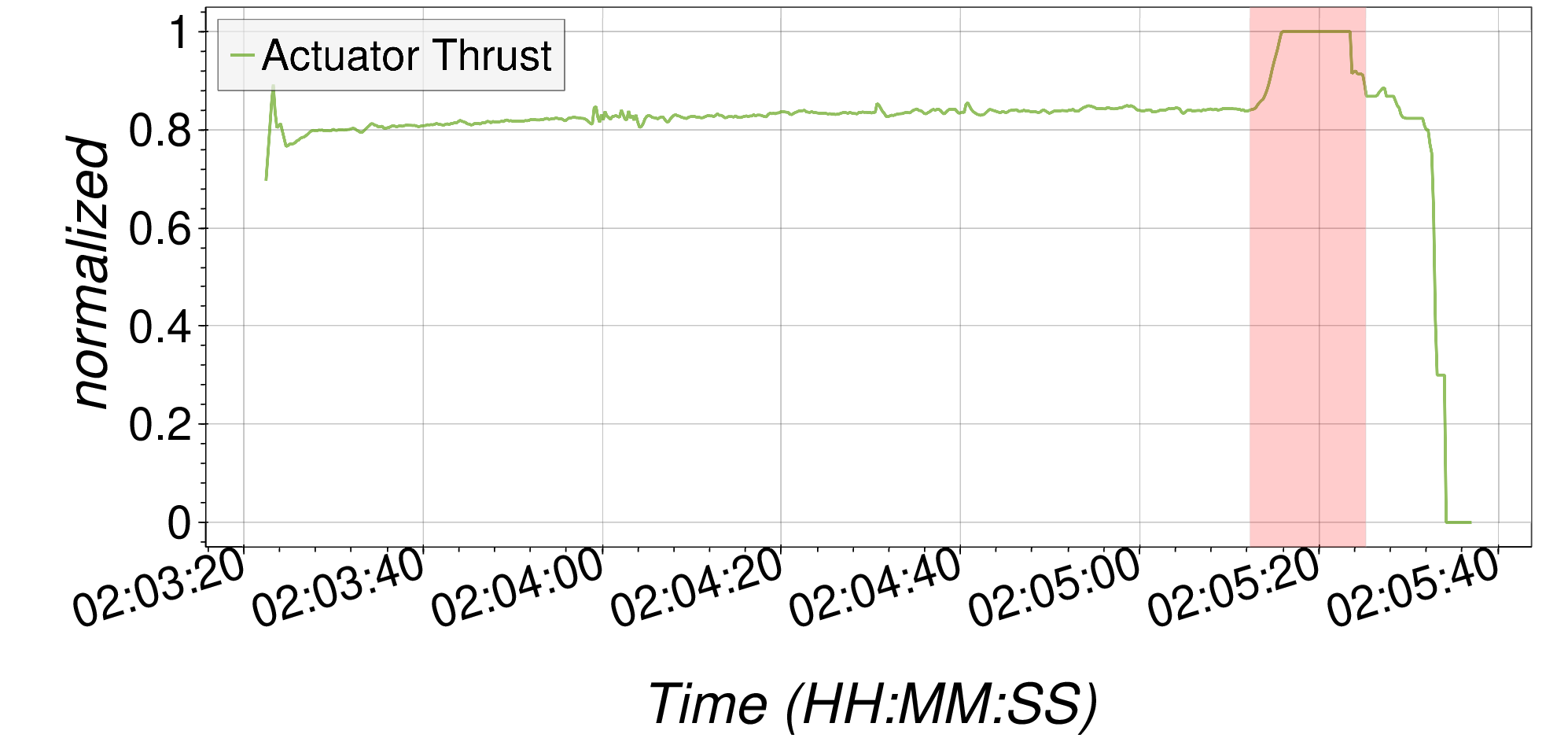}
            \caption{Thrust}
            \label{mechanical_thrust}
        \end{subfigure}

        \caption{Mechanical and Electrical Anomaly Case — Time Series Plots for Position (X, Y, Z), Motor Outputs, Current, Thrust}
        \label{fig:mechanical_combined}
    \end{figure*}

    In the example illustrated in Figure \ref{fig:mechanical_combined}, a combination of thrust saturation and motor output limits, accompanied by current spikes, results in a loss of positional control along the Z axis. However, the Y-axis position remains unaffected; the drift along with X and Z axes can clearly be realized.
    
    Throughout the flight before the annotated duration, the thrust is sufficient to keep the UAV in a desired altitude. However, within the annotated duration, the altitude of the UAV starts to decrease although the thrust is increasing. Thrust cannot compensate altitude decrease and involves positional drift in the X-axis. The relation between thrust, motor outputs, and positions indicates the mechanical and electrical anomaly. This anomalous behavior can be an example of a collective and contextual anomaly.
    
    \subsubsection{External Position Anomalies}

\begin{figure*}
        \centering
    
        \begin{subfigure}[b]{0.48\linewidth}
            \centering
            \includegraphics[width=\linewidth]{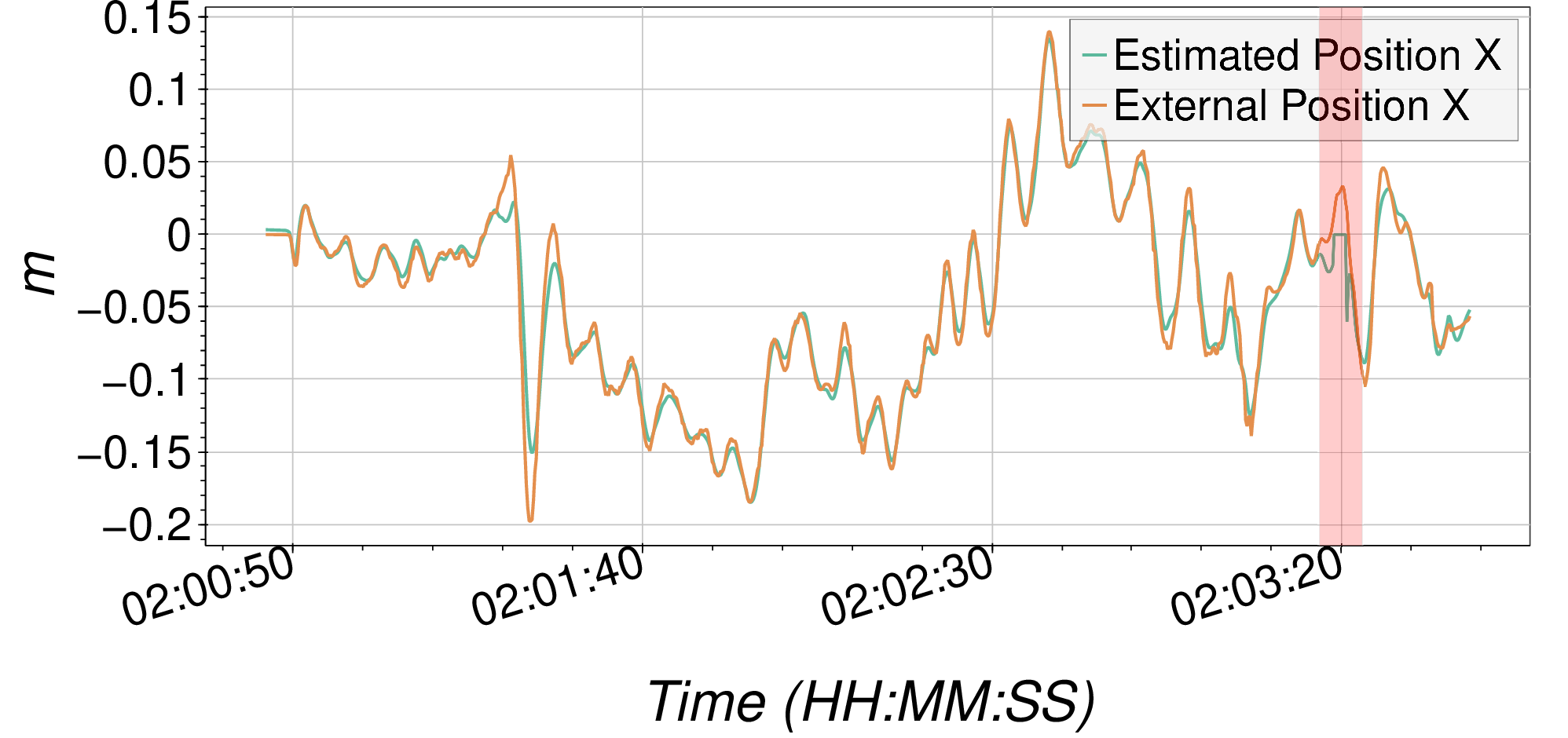}
            \caption{Position X}
            \label{external_position_position_x}
        \end{subfigure}
        \hfill
        \begin{subfigure}[b]{0.48\linewidth}
            \centering
            \includegraphics[width=\linewidth]{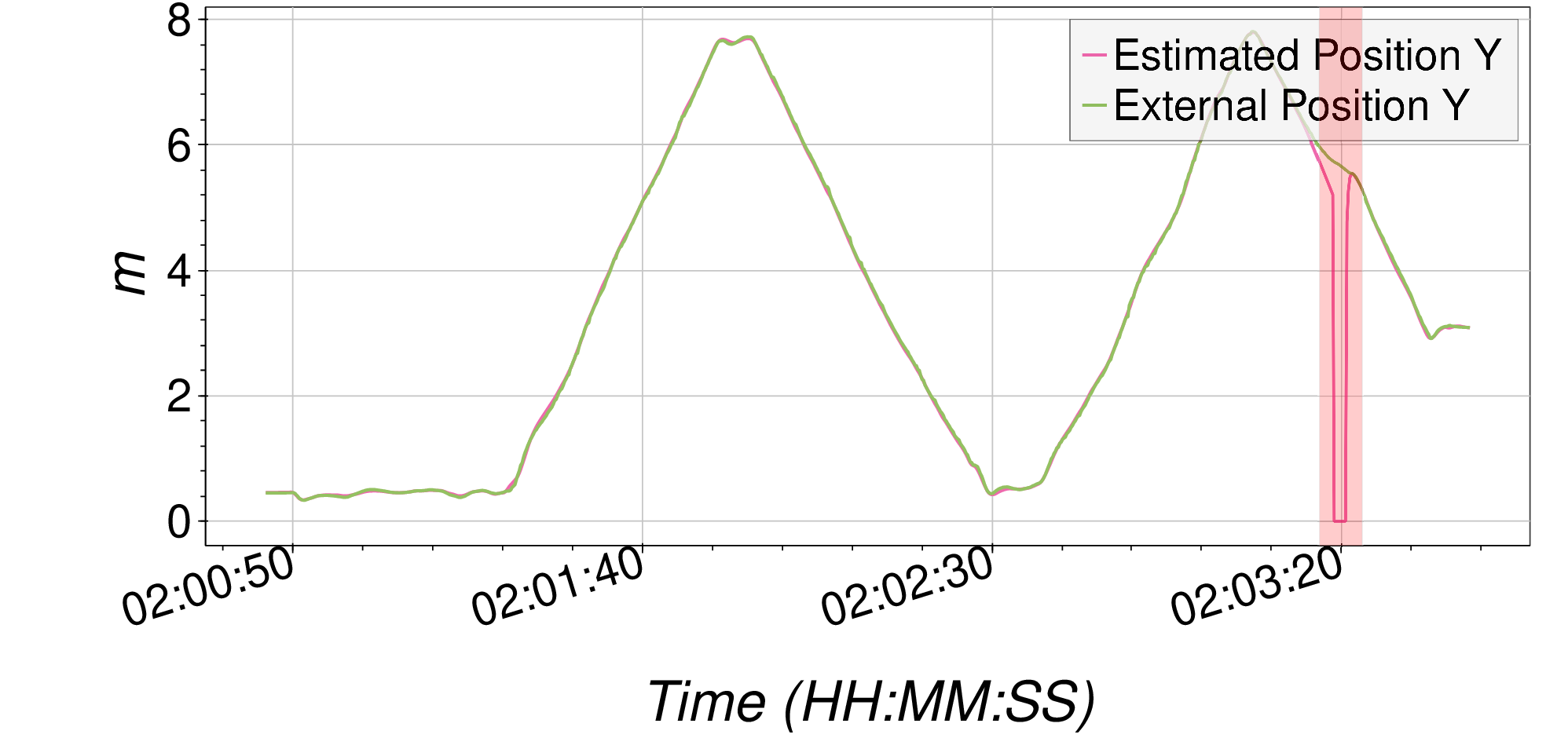}
            \caption{Position Y}
            \label{external_position_position_y}
        \end{subfigure}
    
        \vspace{0.5em}
        
        \begin{subfigure}[b]{0.48\linewidth}
            \centering
            \includegraphics[width=\linewidth]{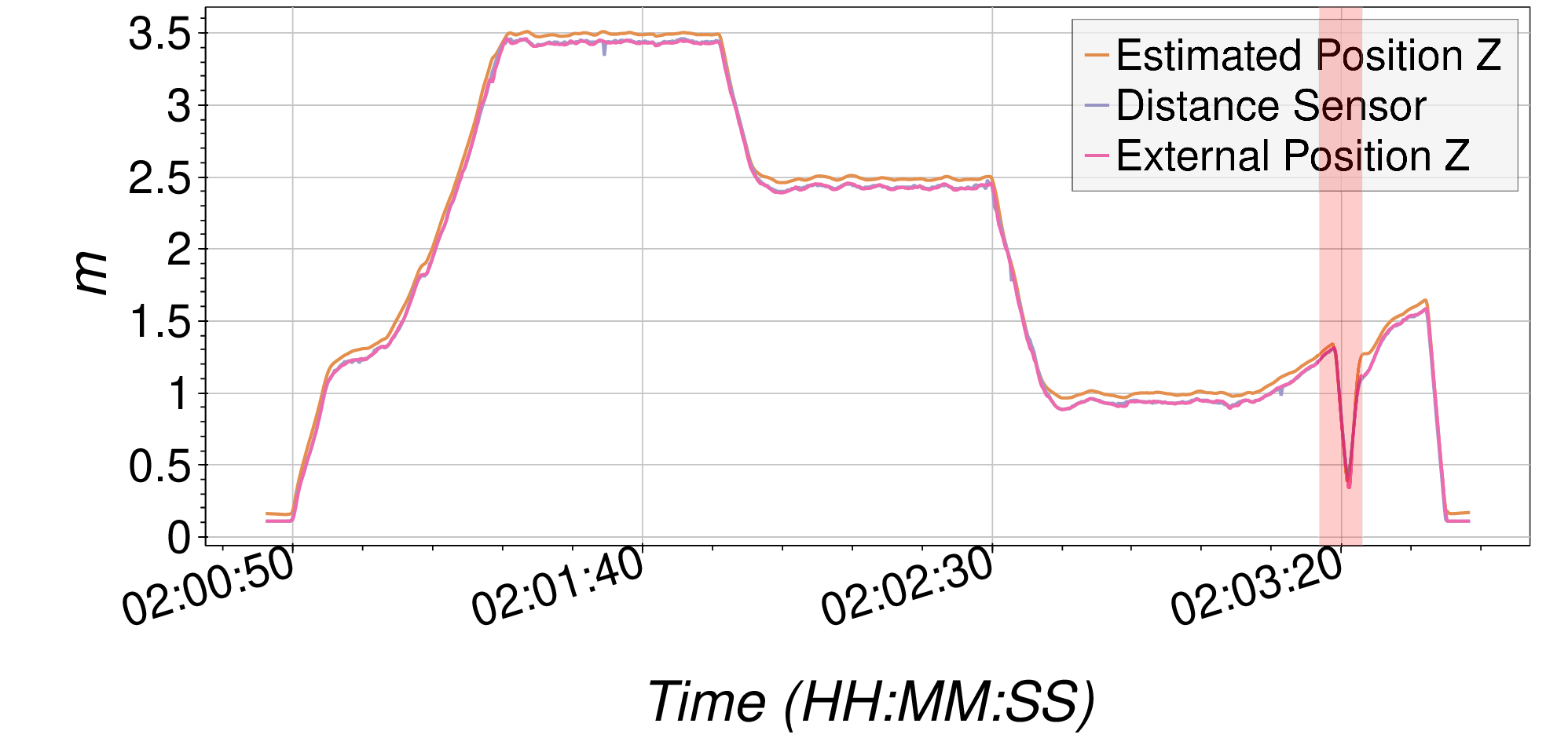}
            \caption{Position Z}
            \label{external_position_position_z}
        \end{subfigure}
        \hfill
        \begin{subfigure}[b]{0.48\linewidth}
            \centering
            \includegraphics[width=\linewidth]{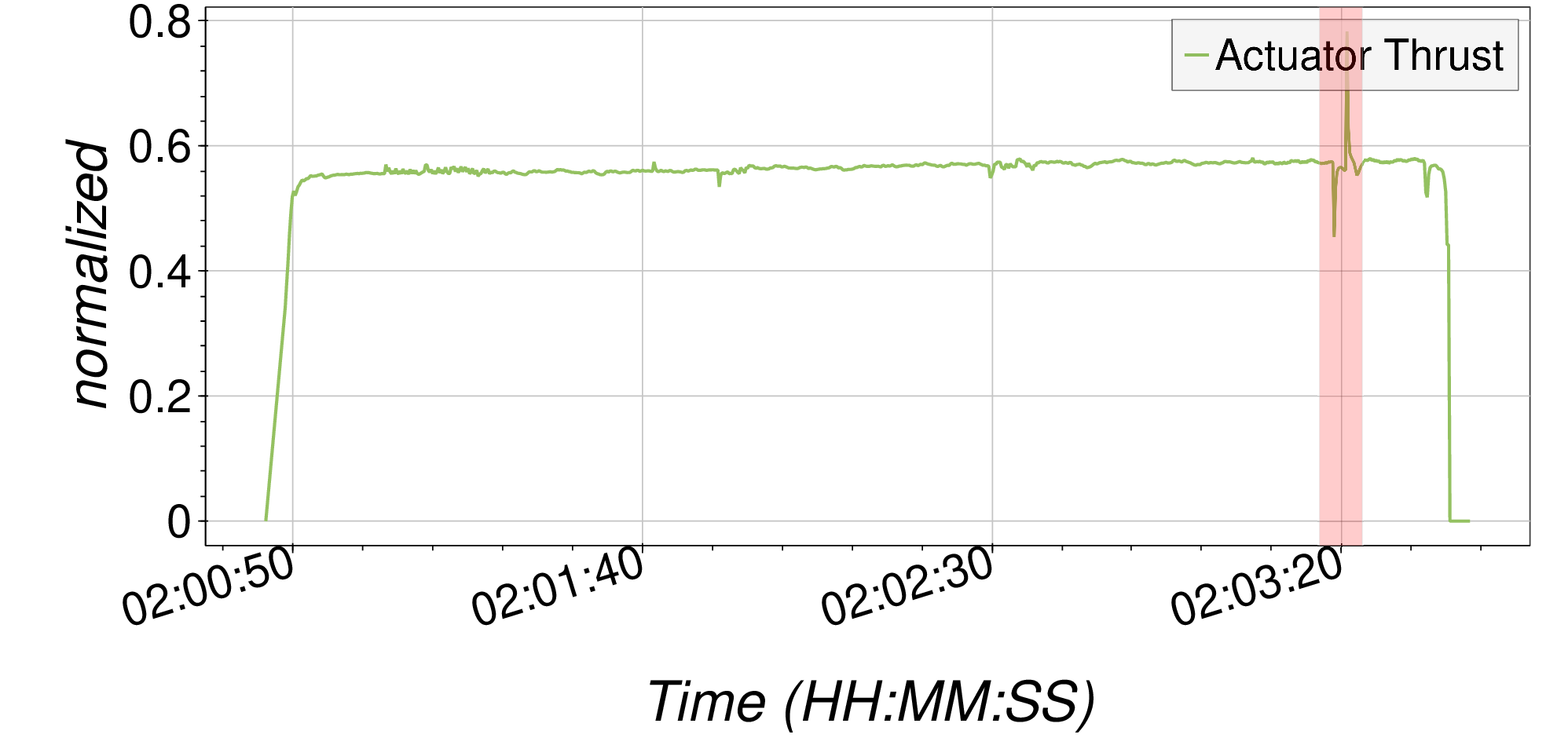}
            \caption{Thrust}
            \label{external_position_thrust}
        \end{subfigure}

        \vspace{0.5em}

        \begin{subfigure}[b]{0.48\linewidth}
            \centering
            \includegraphics[width=\linewidth]{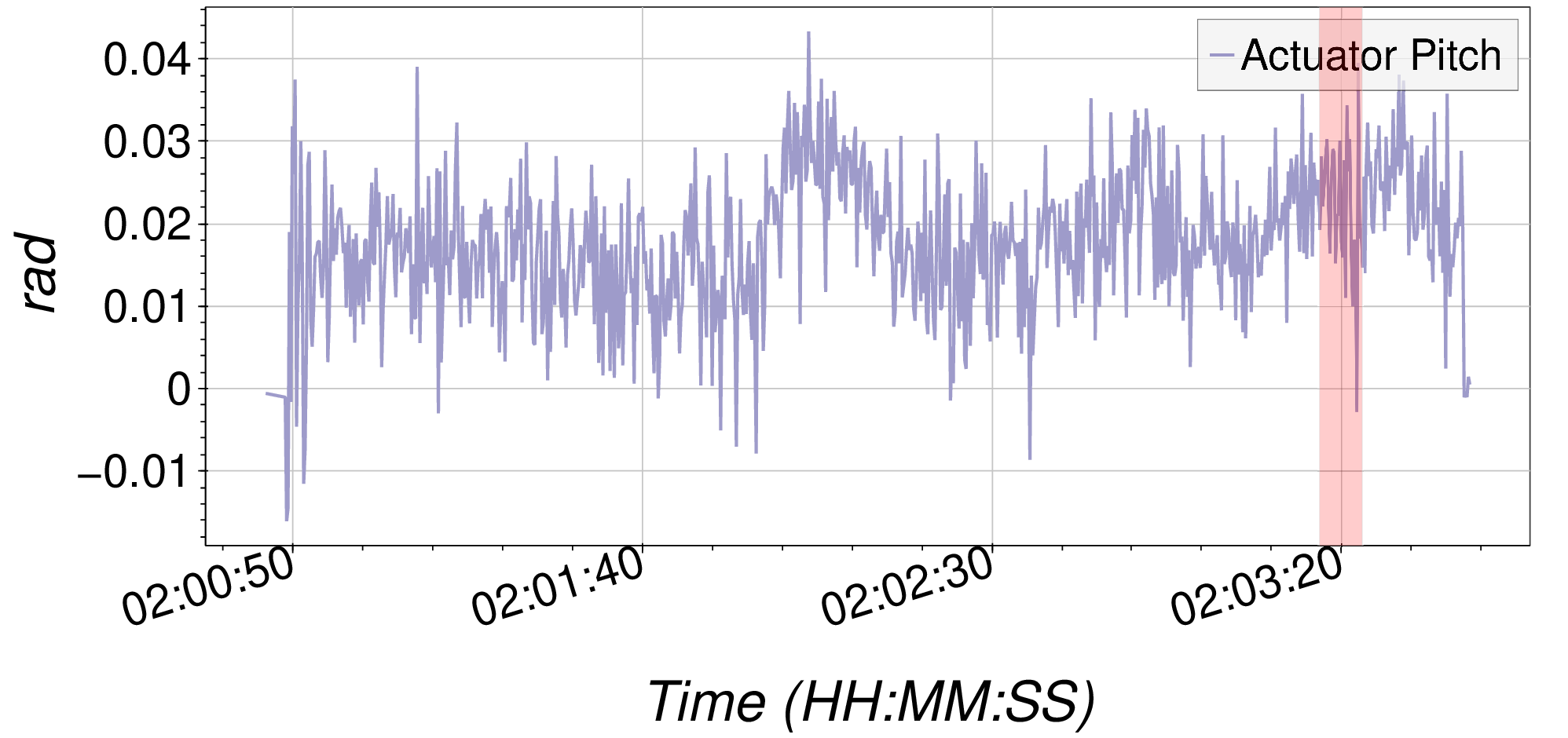}
            \caption{Pitch}
            \label{external_position_pitch}
        \end{subfigure}
        \hfill
        \begin{subfigure}[b]{0.48\linewidth}
            \centering
            \includegraphics[width=\linewidth]{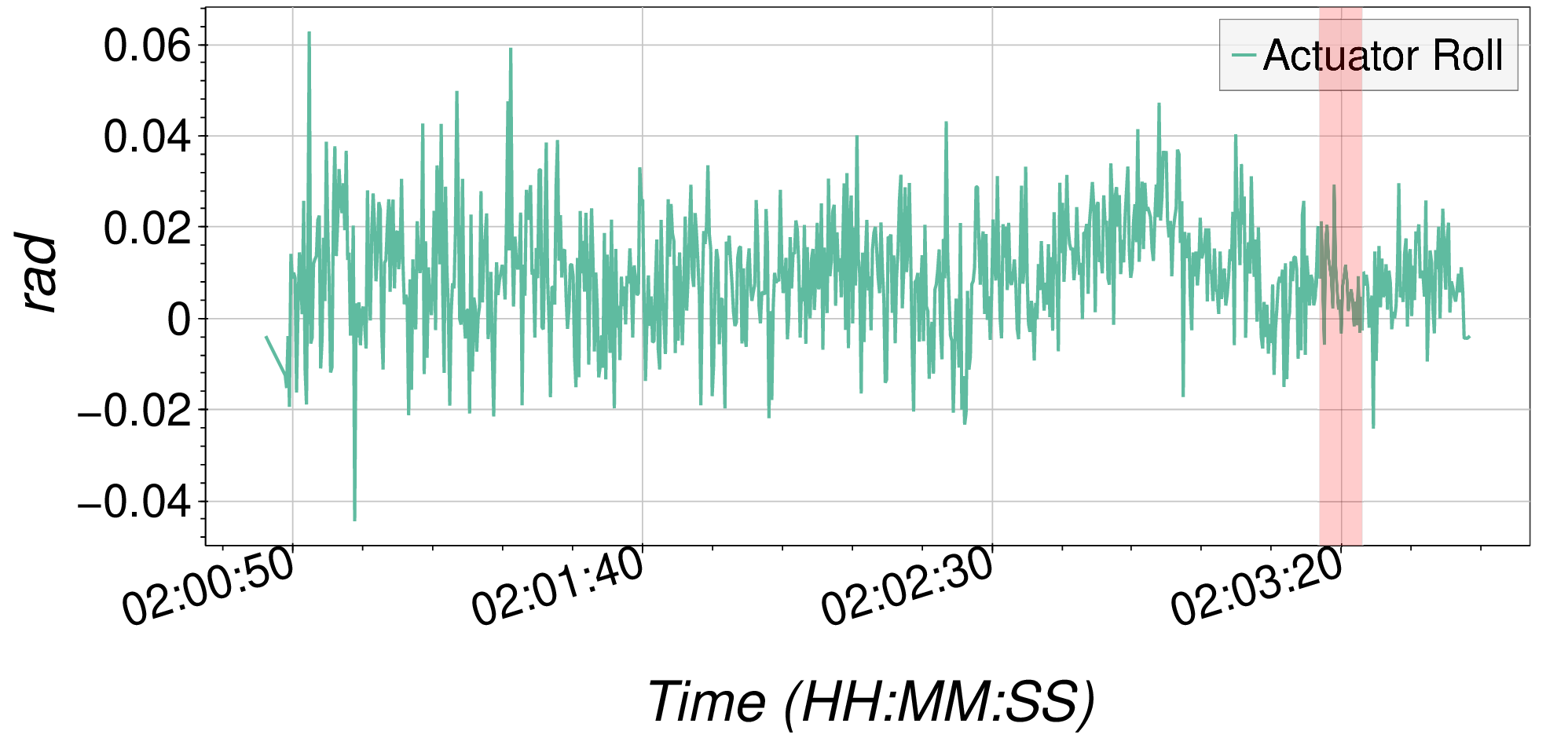}
            \caption{Roll}
            \label{external_position_roll}
        \end{subfigure}

        \caption{External Position Anomaly Case — Time Series Plots for Position (X, Y, Z), Thrust, Pitch, and Roll}
        \label{fig:external_position_combined}
    \end{figure*}

    An illustrative example of an external position anomaly is provided in Figure \ref{fig:external_position_combined}. In these figures, a drift can be observed within the annotated duration of the Y-axis of the position, which is given in Figure \ref{external_position_position_y}, followed by a sudden jump that abruptly resets the estimated position back to the initial state. Notably, throughout both the drift and the jump, there is no significant change in the UAV's roll or pitch angles, suggesting that the attitude was stable during this event. Thrust variation is presented in the same interval; both the distance sensor readings and the altitude estimation appear consistent with this thrust behavior, indicating that the anomaly is not altitude-related.

    Furthermore, the anomaly is isolated to the Y-axis position; no distinctive abnormality is observed in the X-axis position data. The combination of a gradual drift followed by a discrete jump in position Y, without corresponding changes in other relevant flight dynamics, points to a collective and contextual anomaly. Specifically, the estimated Y position abruptly returns to zero during the labeled time range —a behavior that is physically implausible- and contextually inconsistent with the rest of the flight data.

    This anomaly is likely caused by inadequate EKF (Extended Kalman Filter) tuning, where mismatches between sensor noise characteristics, data rates, or covariance settings may result in filter instability. Additionally, possible algorithmic limitations or reset conditions in the EKF implementation may have caused the estimated Y position to be forcibly corrected, manifesting as an external position anomaly.

    \subsubsection{Global Position Anomalies}

\begin{figure*}
        \centering    

        \begin{subfigure}[b]{0.48\linewidth}
            \centering
            \includegraphics[width=\linewidth]{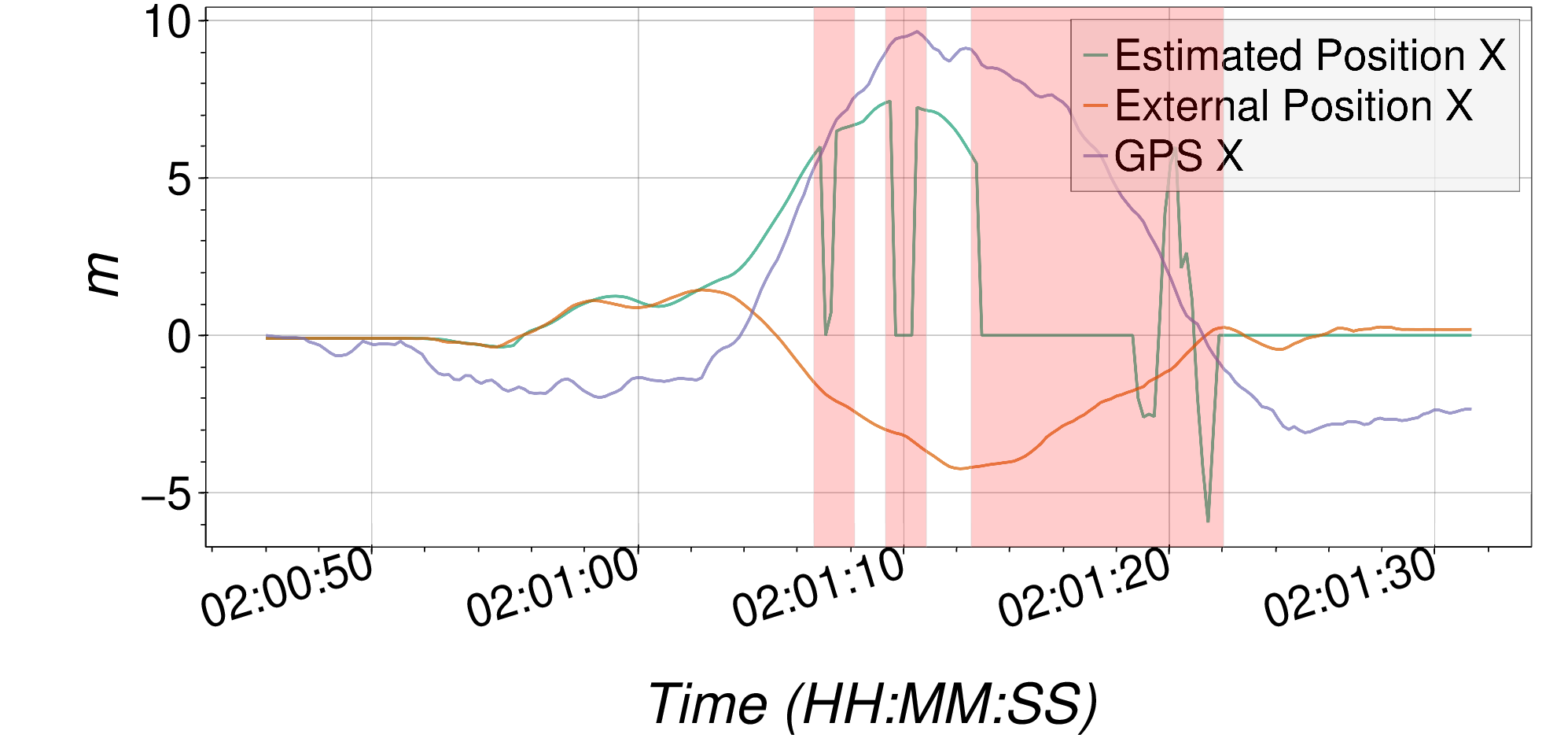}
            \caption{Position X}
            \label{global_position_pose_x}
        \end{subfigure}
        \hfill
        \begin{subfigure}[b]{0.48\linewidth}
            \centering
            \includegraphics[width=\linewidth]{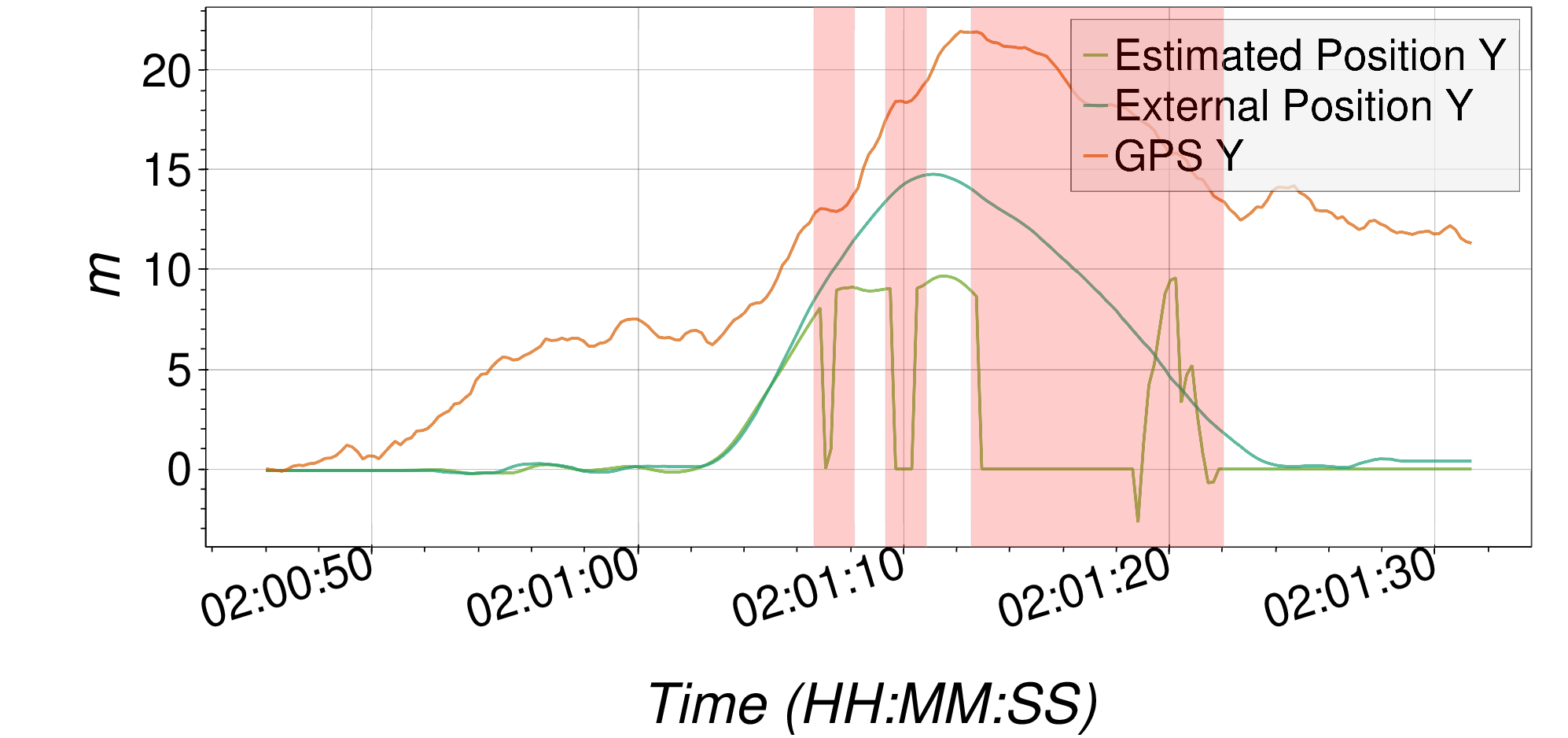}
            \caption{Position Y}
            \label{global_position_pose_y}
        \end{subfigure}

        \vspace{0.5em}

        \begin{subfigure}[b]{0.48\linewidth}
            \centering
            \includegraphics[width=\linewidth]{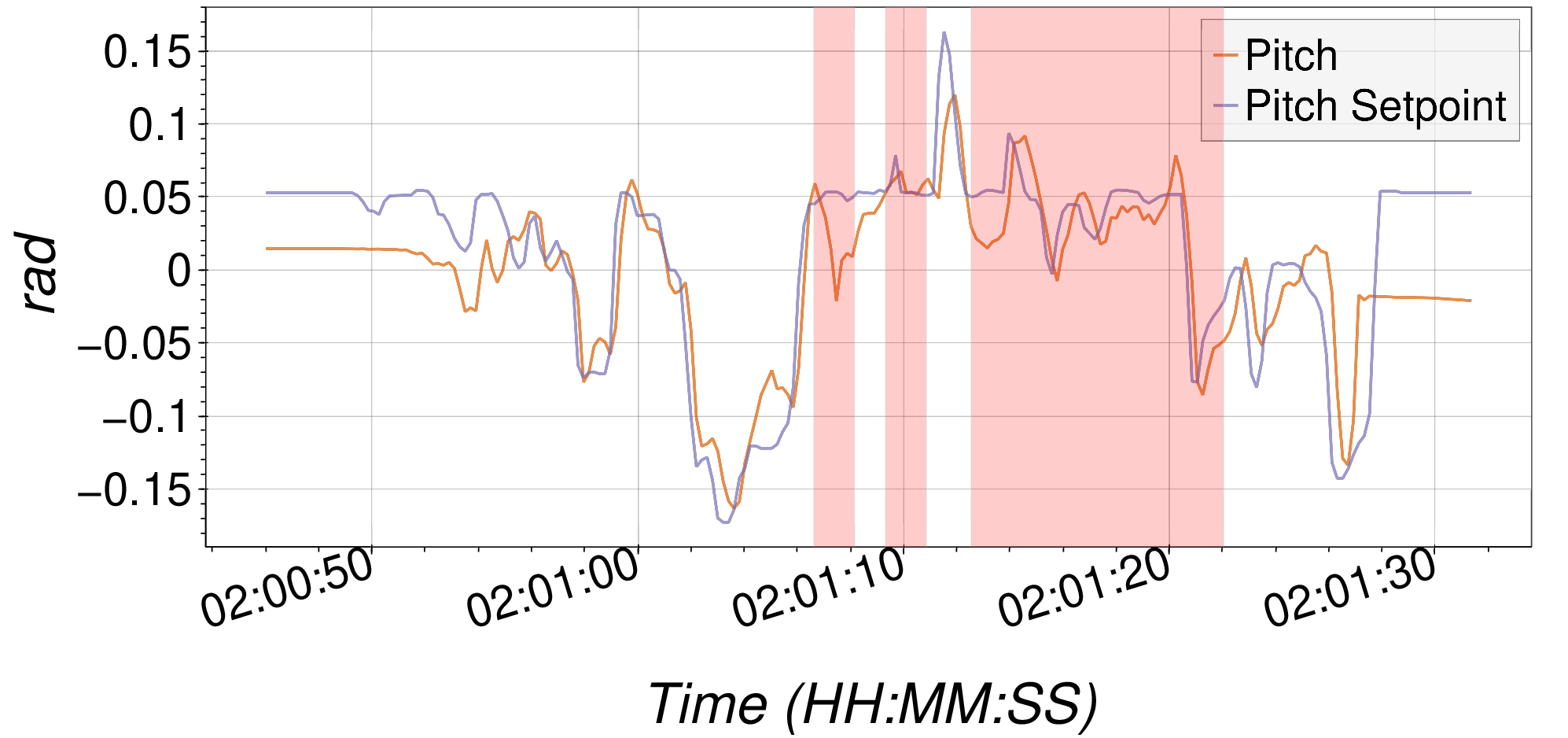}
            \caption{Pitch Setpoint}
            \label{global_position_pitch_setp}
        \end{subfigure}
        \hfill
        \begin{subfigure}[b]{0.48\linewidth}
            \centering
            \includegraphics[width=\linewidth]{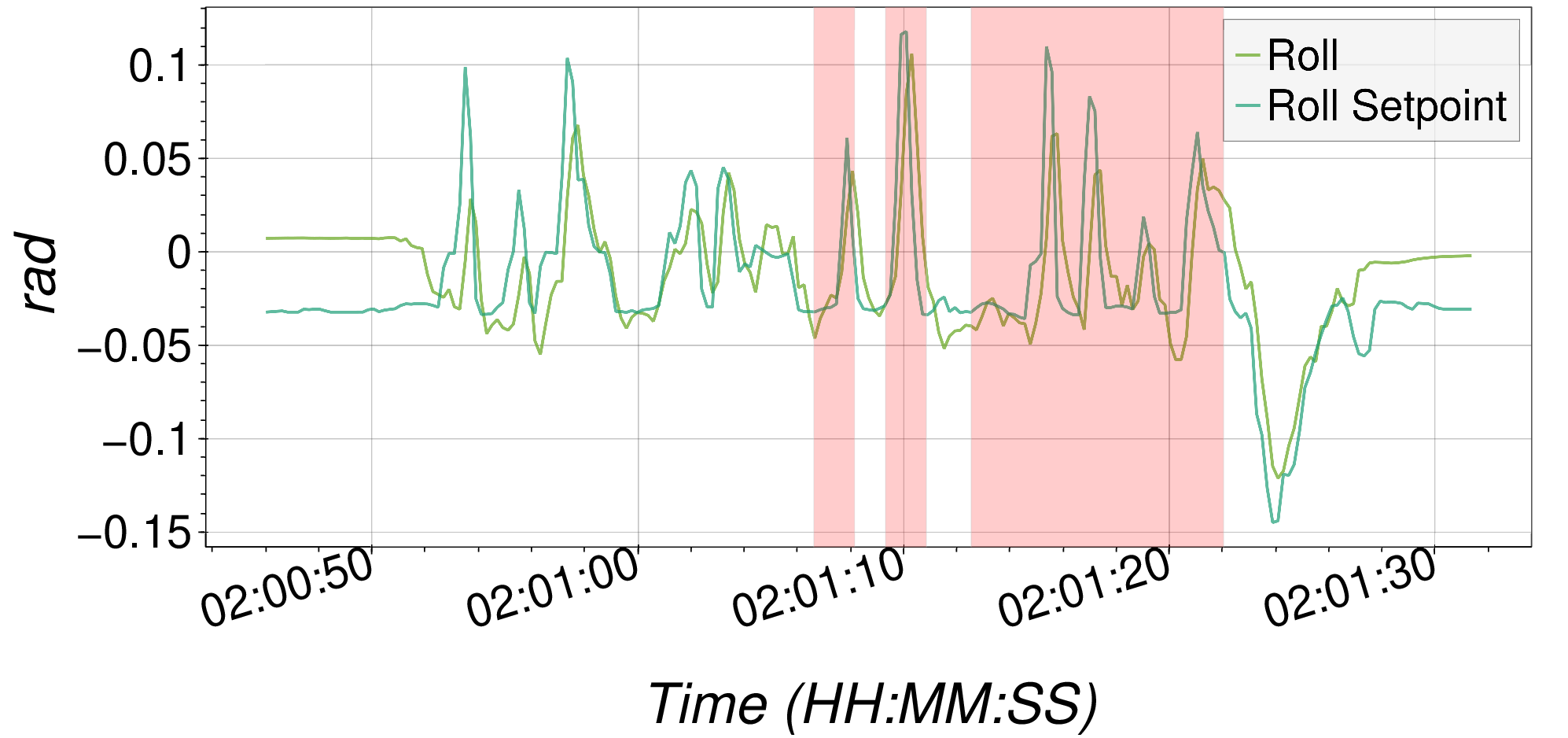}
            \caption{Roll Setpoint}
            \label{global_position_roll_setp}
        \end{subfigure}
    
        \caption{Global Position Anomaly Case — Time Series Plots for Position (X, Y) and Pitch/Roll Setpoints}
        \label{fig:global_position_combined}
    \end{figure*}

    In the example provided in Figure \ref{fig:global_position_combined}, the estimated position, which is not accompanied by corresponding changes in the pitch or roll attitude during the annotated time ranges, despite the pitch and roll changes are minimal and convenient with the autopilot setpoint commands indicating no abnormal maneuver or user input that would justify such positional deviations but the estimated position jumps and is reset a couple of times. A significant mismatch is observed between the estimated position in the X-axis and the external position reference, which reflects both collective and contextual anomaly characteristics. It is collective, because multiple sensor channels must be considered together to understand the inconsistency, and contextual, because the position reset occurs in a flight context where it is against the flight dynamics. Such anomalies often come from issues like GPS signal degradation, fusion inconsistencies, or EKF misalignment.

    \subsubsection{Altitude Anomalies}
    
\begin{figure*}
        \centering
    
        \begin{subfigure}[b]{0.48\linewidth}
            \centering
            \includegraphics[width=\linewidth]{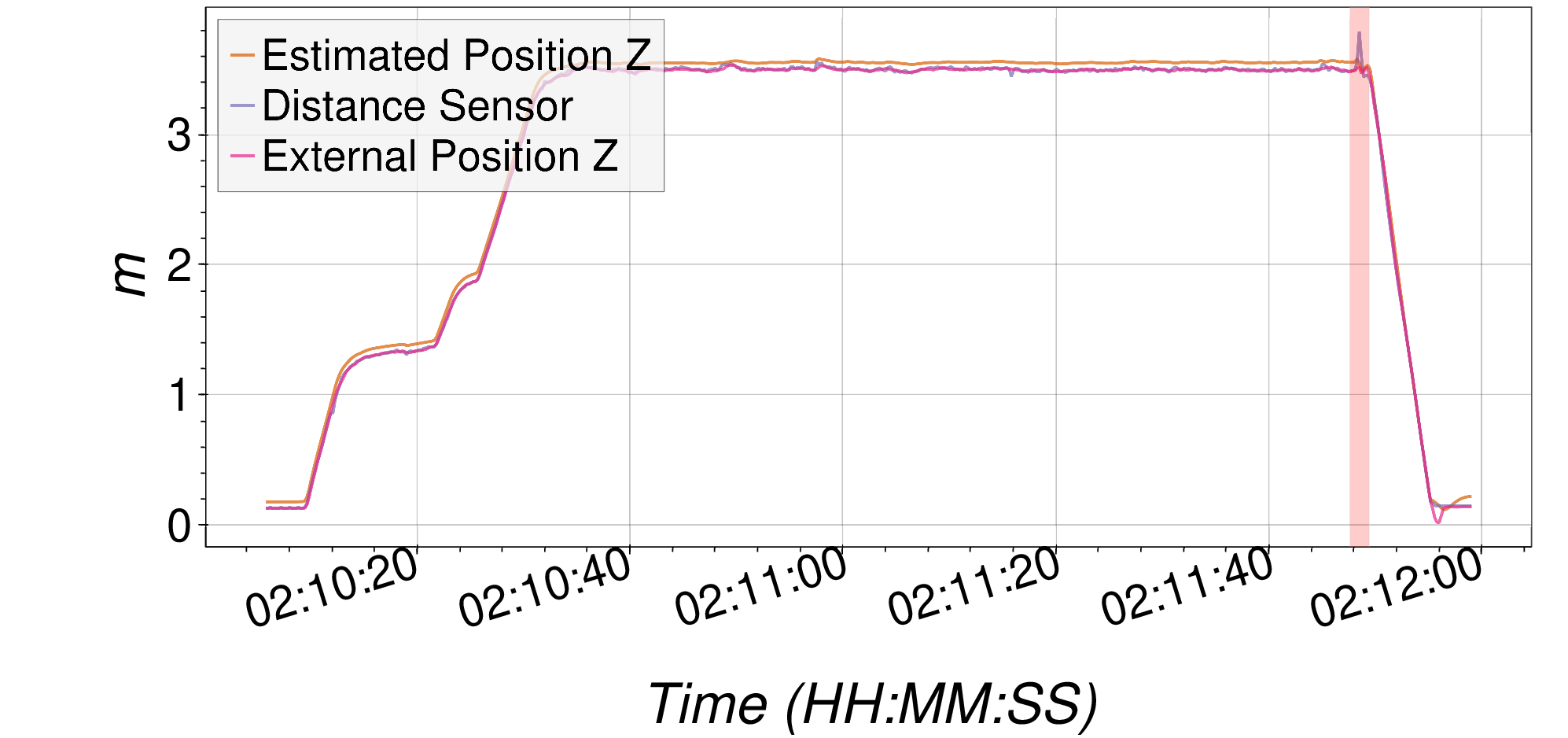}
            \caption{Altitude Z}
            \label{altitude_anomaly_z}
        \end{subfigure}
        \hfill
        \begin{subfigure}[b]{0.48\linewidth}
            \centering
            \includegraphics[width=\linewidth]{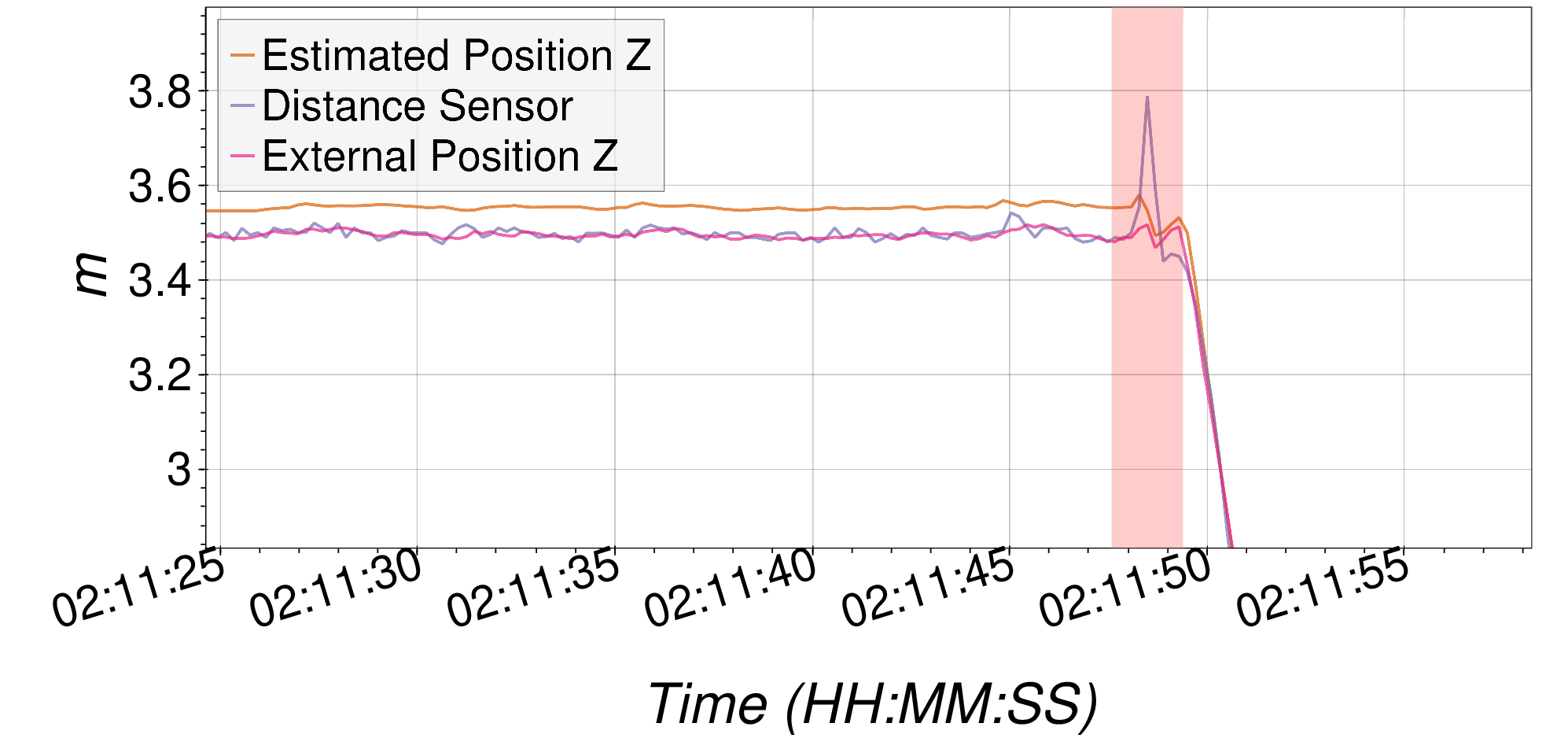}
            \caption{Altitude Z (Scaled)}
            \label{altitude_anomaly_z_scaled}
        \end{subfigure}

        \vspace{0.5em}
        
        \begin{subfigure}[b]{0.48\linewidth}
            \centering
            \includegraphics[width=\linewidth]{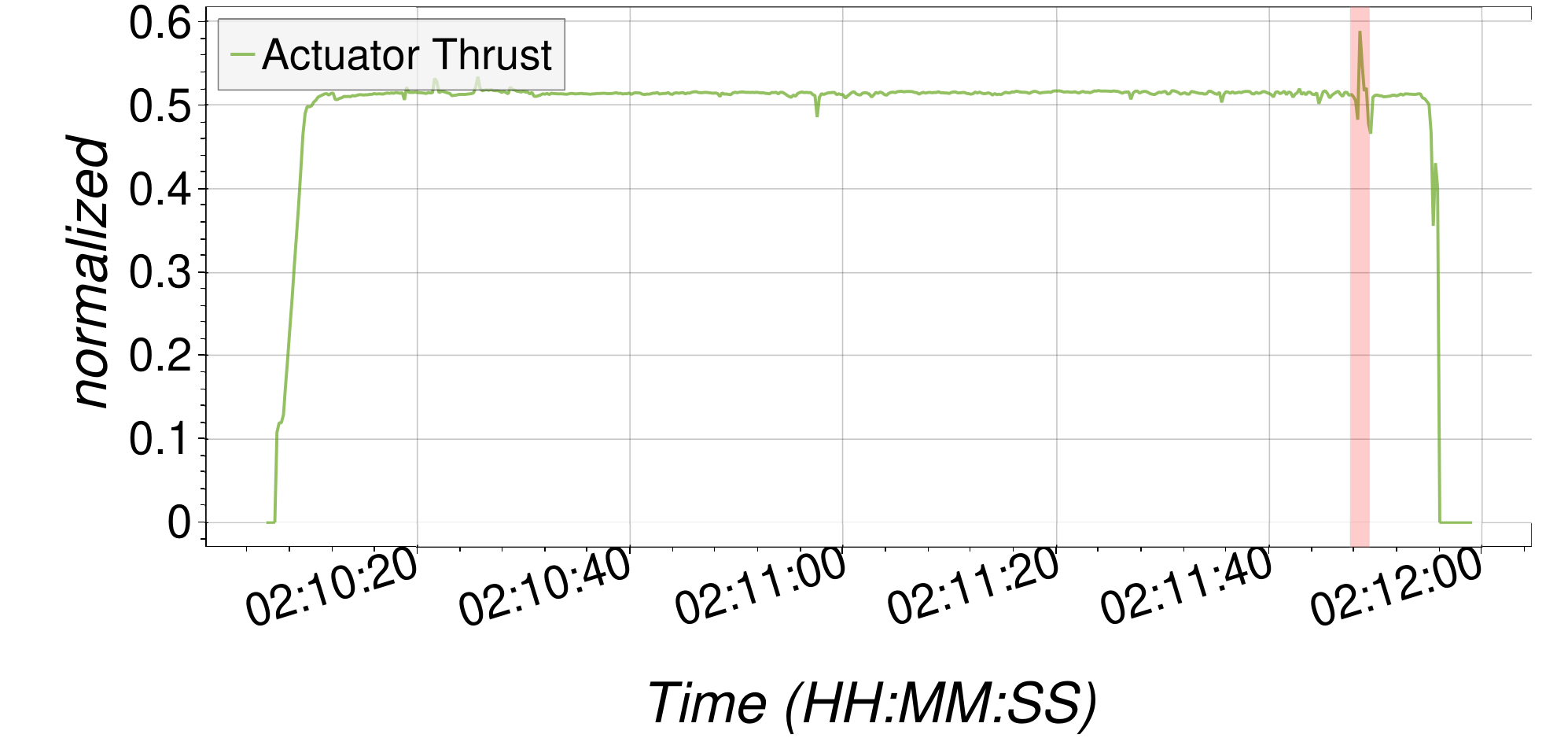}
            \caption{Thrust}
            \label{altitude_anomaly_thrust}
        \end{subfigure}
        \hfill
        \begin{subfigure}[b]{0.48\linewidth}
            \centering
            \includegraphics[width=\linewidth]{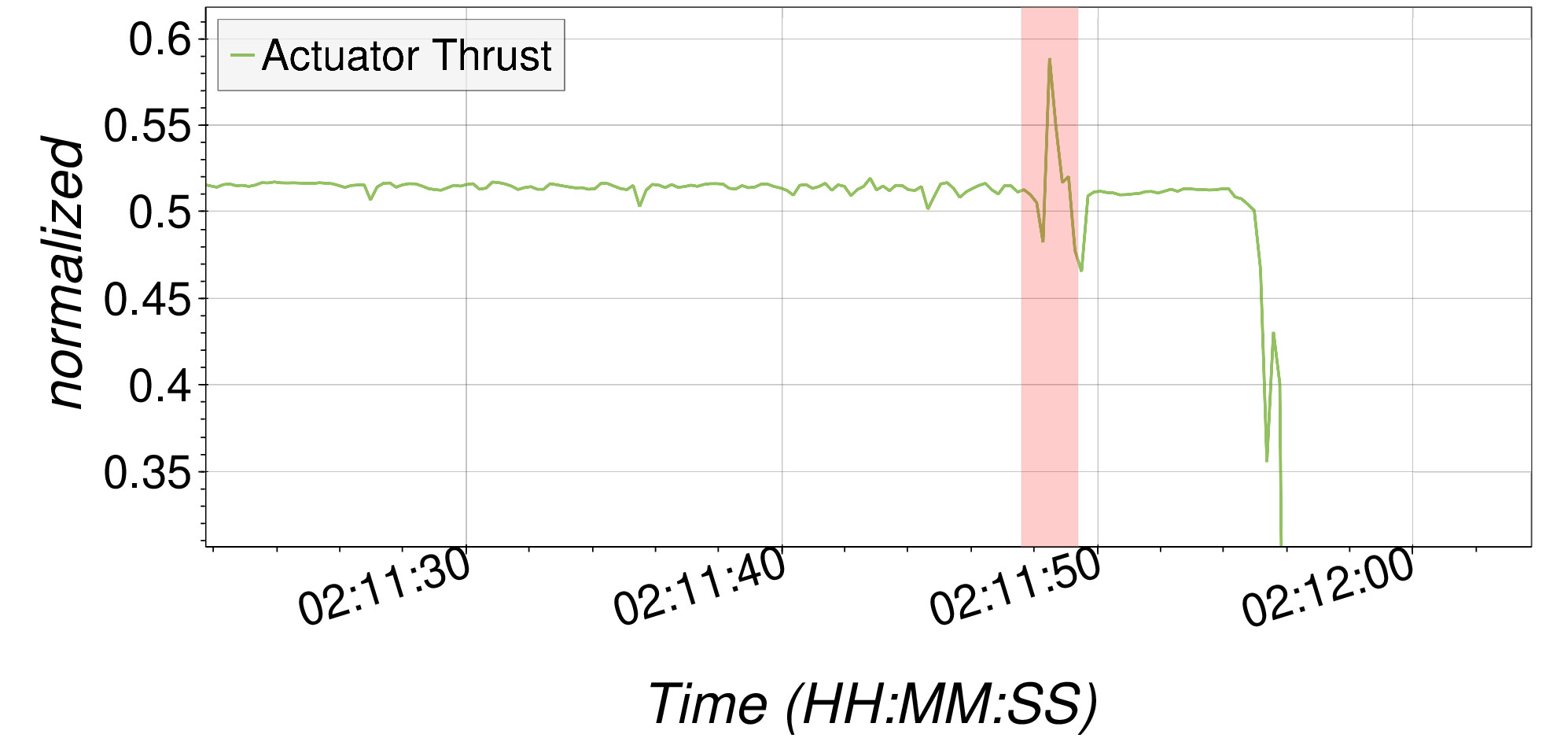}
            \caption{Thrust (Scaled)}
            \label{altitude_anomaly_thrust_scaled}
        \end{subfigure}
    
        \caption{Altitude Anomaly Case — Time Series Plots for Position (Z) and Thrust}
        \label{fig:altitude_anomaly_combined}
    \end{figure*}

    Figure \ref{fig:altitude_anomaly_combined} illustrates an example of an altitude anomaly. In this example, the anomaly becomes apparent by observing the correlation between the thrust output, distance sensor measurements, and the estimated local Z position. As shown in Figure \ref{altitude_anomaly_thrust}, the autopilot issues abrupt thrust changes—alternating between increased and decreased thrust. This occurs simultaneously with notable changes in the distance sensor measurements, indicating that the UAV's relative altitude is indeed changing. Despite this, the estimated local Z position remains unchanged. This discrepancy occurs because the Extended Kalman Filter (EKF) disregards the sudden altitude change as an outlier, which is consistent with the Bayesian filtering approach used to suppress noise. However, in this case, the altitude change is not merely noise—it is corroborated by thrust commands and sensor readings. Thus, the EKF’s suppression results in an inaccurate altitude estimate.

    This type of anomaly is best described as a collective anomaly, not a simple outlier. It is only detectable through semantic analysis of the UAV's behavior, highlighting the importance of considering multiple sensor inputs in combination when diagnosing the estimation anomalies.

\section{Review and Comparison of UAV Anomaly Datasets} \label{comparison_of_datasets}

\begin{table*}
\caption{Comparison of Flight Anomaly Datasets}{N/A: No data is available in the paper, ** Total of simulation and real flights.}
\centering
\resizebox{\linewidth}{!}{
\renewcommand{\arraystretch}{1.3} 
\begin{tabular}{|l|cc|cc|cc|c|c|c|}
\hline
\textbf{Dataset Name} & \textbf{Dataset Focus} & \textbf{Aircraft Type} & \textbf{\makecell{\# Real \\ Flights}} & \textbf{\makecell{Real Flight \\ Duration $(\sim)$}} & \textbf{\makecell{\# Simulated \\ Flights}} & \textbf{\makecell{Simulated Flight \\ Duration $(\sim)$}} & \textbf{Control Method} & \textbf{Autopilot} & \textbf{Environment} \\ \hline
ALFA \cite{keipour2021alfa} & Motor \& Actuator Faults & Fixed wing & 47 & 79 min & - & - & Autonomous & Ardupilot & Outdoor \\ 
UAV-FD \cite{baldini2023uav} & Actuator Faults (chipped blade) & Hexarotor & 18 & N/A & - & - & Manual & Ardupilot & Outdoor \\ 
PADRE \cite{puchalski2023padre} & Propeller Failures & Quadrotor & 20 & 57 min & - & - & Manual & Parrot FC & Indoor \\ 
BASiC \cite{ahmad2024uav} & Sensor Faults & Quadrotor & - & - & 70 & 441 min & Autonomous & Ardupilot & Simulation \\ 
RflyMAD \cite{le2023rflymad} & Actuator \& Sensors Faults & Quadrotor & 497 & N/A & 5132 & **3283 min & Autonomous & PX4 & Simulation + Outdoor \\ 
CrazyPAD \cite{masalimov2024crazypad} & Propeller Anomalies & Quadrotor & 224 & 100 min & - & - & Autonomous & Crazyflie & Indoor \\ 
Blackbird \cite{antonini2020blackbird} & VI-SLAM Errors & Quadrotor & 168 & 617 min & - & - & Autonomous & Custom & Indoor \\
Euroc \cite{burri2016euroc} & Sensor fusion benchmarking & Hexarotor & 11 & 21 min & - & - & Autonomous + Manual & AscTec autopilot & Indoor \\
UAS-Structure \cite{gururajan2019flights} & Propeller Failures & Hexarotor & 20 & 100 min & - & - & Manual + Autonomous & Ardupilot & Enclosed Outdoor \\
TLM:UAV \cite{yang2023acquisition} & GPS,IMU,Motor,RC & Quadcopter & - & - & 4 & 20 min & Autonomous & Ardupilot & Simulation \\
\textbf{UAV-SEAD} (Ours)& State Estimation Anomalies & Quadrotor & \textbf{1396} & \textbf{3140 min } & - & - & Autonomous + Manual & PX4 & Indoor + Outdoor \\ \hline
\end{tabular}
}
\label{tab:anomaly_data_separate_columns}
\end{table*}

Anomaly detection and identification using a data-driven approach requires the collection of large amounts of data, and creating anomalous or faulty cases is a challenging task, especially for aerial vehicles. Several datasets have been presented to advance the studies conducted in this field to provide opportunities for developing and testing new data-driven anomaly detection and isolation methodologies. In this paper, a dataset is presented for state estimation anomalies, and this is a new classification approach for UAV anomalies. Therefore, the UAV datasets that include any anomaly are presented for comparison to show the data source of the anomaly detection research area.

The most commonly used UAV anomaly dataset in the literature so far is ALFA \cite{keipour2021alfa}, which addresses actuator and motor anomalies in fixed-wing aircrafts, including full motor failure and stuck rudder, aileron, and elevator faults. It contains a total of 47 flights, with a combined duration of 79 minutes. The ALFA dataset provides real flight data from fixed-wing UAVs, encompassing 23 full engine failure scenarios and 24 actuator fault scenarios. It includes both raw and processed data, along with ground truth annotations for fault types and timings.

UAV-FD \cite{baldini2023uav} is a dataset that includes actuation failures caused by chipped blades. The dataset includes flights collected on a Hexarotor DJI frame equipped with Pixhawk Ardupilot. It consists of 12 faulty flights out of 18 manual outdoor flights, with the chipping of the blades injected to examine various effects.

PADRE is another dataset involving \cite{puchalski2023padre} propeller anomalies, such as chipped edges and bent tips on blades. These structural damages lead to propeller failures, which in turn affect the actuators and can be classified as actuator anomalies. The dataset consists of 20 manual flights conducted with a Parrot Bebop 2 quadcopter. Only gyroscope and accelerometer data are reported to detect the anomalies.

Another dataset, BASiC (Biomisa Arducopter Sensory Critique) \cite{ahmad2024uav}, categorizes anomalies into six distinct sensor failure types: global positioning system (GPS), remote control, accelerometer, gyroscope, compass, and barometer. It includes 70 autonomous flights totaling over 7 hours, as 3 hours of pre-failure data, 3 hours of post-failure data, and approximately 1 hour of flight data without any failures. It includes various sensor failure scenarios, providing a rich resource for analyzing UAV sensor anomalies.

RflyMAD \cite{le2023rflymad} is the largest dataset in the comparison, with 114 GB of data including 11 types of faults across 6 flight statuses to account for various scenarios in which multicopters exhibit different mobility levels during faults. The dataset consists of both simulation data based on RflySim and real flight data, with each case containing ULog and telemetry logs. It considers actuators (motors, propellers), sensors (accelerometer, gyroscope, magnetometer, barometer, and GPS), and environmental effects. The dataset contains a total of 5,629 flight cases, with 3,283 minutes of fault time. It includes 2,566 cases from Software-in-the-Loop (SIL) simulation, 2,566 cases from Hardware-in-the-Loop (HIL) simulation, and 497 cases from real flights.  A small model is applied to the dataset to demonstrate the dataset's performance.

Crazyflie Propeller Anomaly Dataset \cite{masalimov2024crazypad} addresses structural damage in nano quadcopters, such as impaired propellers or asymmetrical weight distribution, and examines their effects on flight patterns, power consumption, stability, and control responsiveness. The dataset includes case studies that demonstrate its effectiveness using data-driven methods.

The Blackbird dataset \cite{antonini2020blackbird} is distinctive among other referenced datasets as it addresses the VI-SLAM problem for high-speed racing drones in indoor environments, generating virtual images from real flight records. Most datasets and research on fault detection and isolation (FDI) systems for UAVs focus on sensor and actuator failures, as well as blade anomalies. However, Antonini et al. \cite{antonini2020blackbird} concentrate on state estimation for racing drones, specifically considering the output from visual-inertial SLAM. While the Blackbird dataset provides camera images for visual inertial SLAM, it can be considered as a state generation database because VI-SLAM algorithms will produce position/orientation as a state; besides, ground-truth motion capture position outputs are provided along with it. It provides data from a forward stereo camera and a downward camera for 168 flights, totaling more than 10 hours of flight duration, which are rendered photorealistically using FlightGoggles.

The EuRoC Micro Aerial Vehicle Datasets \cite{burri2016euroc} are primarily focused on SLAM and VIO (Visual Inertial Odometry) benchmarking, but may contain valuable inertial and visual-inertial drift anomalies. The dataset is one of the most widely used public datasets for evaluating visual-inertial odometry and SLAM systems. It consists of real-world flight data collected using an AscTec Firefly hex-rotor platform equipped with synchronized stereo cameras and an IMU. The dataset includes 11 indoor flight sequences across various environments (e.g., machine hall, Vicon room) with varying difficulty levels. Although not specifically designed for anomaly detection, the high-precision sensor logs and diverse flight conditions make it a valuable benchmark for state estimation research and testing anomaly detection algorithms under realistic scenarios. The flights were executed in semi-autonomous and manual modes. 

Structural anomalies due to damaged propellers are investigated in \cite{gururajan2019flights}, which provides real-world UAV flight data collected under intentionally injected structural faults on composite propellers. The dataset is designed to analyze the effects of blade damage and failures on multirotor flight performance. The aircraft used is a hexarotor UAV flown in a controlled environment with progressive damage levels applied to the propeller blades. The dataset includes real flight logs annotated with the severity of each fault condition and flight outcome. This dataset is especially valuable for fault detection, isolation, and recovery (FDIR) studies, as it captures the system’s response under degraded physical conditions. The flights were conducted using a Pixhawk autopilot with ArduPilot firmware, with manual or semi-autonomous flight control.

Yang et al. \cite{yang2023acquisition}  present a Time Line Modeling (TLM) scheme that runs a SITL-based ArduPilot simulation to inject four representative faults as GPS, accelerometer, engine, and remote control(RC) failures within the same mission. The Time Line Modeling approach is structured into two distinct stages. In the first stage, normal and abnormal segments within the UAV flight data are identified using a Fault Point-in-Time Anchoring method. The second stage extends the identified abnormal segments through a Time-Window Stretching technique to balance the dataset by increasing the proportion of abnormal samples. Prior to model training, time-related and non-generic features are removed from the dataset. The refined data is then used to train and evaluate classical machine learning algorithms for anomaly detection.


Table \ref{tab:anomaly_data_separate_columns} presents a comparison of the datasets discussed above. In the aforementioned datasets, anomaly cases are manually injected (e.g., by cutting propellers, spoofing sensors, or triggering software failures).

Most studies rely on limited real flight data, often focusing on hardware anomalies/failures, and some of them usees only simulation environments. They typically break hardware or insert anomaly-generation modules into the flight control software for testing purposes, which may not accurately reflect real-world conditions. Some studies employ custom logging mechanisms specific to their flight controllers or system designs, but there are two de-facto standards: PX4 and Ardupilot. Due to community-driven improvements, PX4 has become the most commonly used flight controller in recent research. 

None of the given datasets consists solely of naturally occurring faults. Although the EuRoC MAV dataset, which is designed for visual-inertial SLAM, includes purely real flights in both a machine hall and a Vicon room, no faults or anomalies are present. However, it is included here because the data can be used for state estimation research. In the Blackbird dataset, all flights are healthy aggressive maneuvers; the vision streams are rendered afterward via FlightGoggles. Again, there are no real-world faults at all.

In this study, we present UAV-SEAD, a dataset including a large amount of real flight data using PX4 and its logging standard, which was collected, focusing on state estimation anomalies rather than presenting failures in fundamental aircraft components like wings, rotors, or actuators. The dataset contains no injected anomalies; all reported anomalies occur naturally during real flights.


\section{Conclusion and Future Works} \label{conclusion_and_future_works}
This paper introduces a comprehensive dataset, UAV-SEAD, designed to aid in the development of state estimation anomaly detection systems for UAVs. 
This dataset aims to address the limitations of existing UAV anomaly datasets, offering a larger volume of real flight data, a broader range of anomaly types, and a focus on state estimation anomalies. It provides a detailed classification of state estimation anomalies into categories such as mechanical and electrical, external position, global position, and altitude anomalies. This classification, along with the provision of labeled data and annotation tools, supports the development and evaluation of effective anomaly detection methodologies.

A key direction for future work involves the development of a data-driven model specifically tailored for state estimation anomaly detection in UAV systems trained on UAV-SEAD. This model will be designed to process the multivariate time-series data from the aforementioned UAV data streams presented in the dataset, to accurately identify deviations from normal flight behavior. It is required to support real-time monitoring of UAV operations, providing immediate alerts upon the detection of any anomalous behavior by enabling timely intervention to prevent potential failures for ensuring reliable and safe operation.


\section*{Acknowledgment}
The authors would like to thank Mehmet Boztepe for his efforts and valuable contributions to all the flight tests, especially his RC pilot skills and engineering expertise.


\clearpage
\bibliographystyle{IEEEtran}
\bibliography{IEEEabrv,referencelist}

\end{document}